\pdfoutput=1

\documentclass[11pt]{article}

\usepackage{EMNLP2023}

\usepackage{times}
\usepackage{latexsym}
\usepackage{amsmath}

\usepackage{sidecap}

\usepackage[textsize=footnotesize]{todonotes}

\newcommand{\myvec}[1]{\mathbf{#1}}
\newcommand{\mymat}[1]{\mathbf{#1}}

\usepackage{tikz}
\usetikzlibrary{shadows}

\usepackage{mathtools}
\mathtoolsset{showonlyrefs,showmanualtags}

\newcommand{\rstack}[4][3cm]{\tikz[baseline]{
    \foreach \k[count=\n] in {#2}{\xdef\numrecs{\n}}
    \foreach \k[count=\n, evaluate=\n as \c using \n/\numrecs*100] in {#2}{
        \node[shift={(-\n/7,-\n/7)}, draw, thick, fill=#3!\c, drop shadow, inner sep=0, minimum height=\k cm, minimum width=#1, anchor=north east](r\n) at (0,0){};
    }
    \node at (r\numrecs){#4};
}}

\usepackage[T1]{fontenc}

\usepackage[utf8]{inputenc}

\usepackage{microtype}

\usepackage{inconsolata}

\usepackage{graphicx}
\usepackage{amsmath,amsfonts,amssymb}
\usepackage{subcaption}
\usepackage{array}
\usepackage{makecell}
\usepackage{multirow}
\usepackage{multicol}
\usepackage{mathtools}
\usepackage{inconsolata}

\newenvironment{itemizesquish}[2]{\begin{list}{\labelitemi}{\setlength{\itemsep}{#1}\setlength{\labelwidth}{#2}\setlength{\leftmargin}{\labelwidth}\addtolength{\leftmargin}{\labelsep}}}{\end{list}}

\title{A Joint Matrix Factorization Analysis of Multilingual Representations}

\author{Zheng Zhao \quad Yftah Ziser \quad Bonnie Webber \quad Shay B. Cohen \\
  Institute for Language, Cognition and Computation \\
  School of Informatics, University of Edinburgh \\
  10 Crichton Street, Edinburgh, EH8 9AB \\
  \texttt{\{zheng.zhao,yftah.ziser,bonnie.webber\}@ed.ac.uk},
  \texttt{scohen@inf.ed.ac.uk}}

\newcommand{\rv}[1]{\mathbf{#1}}

\newcommand{\elementrv}[1]{\textnormal{#1}}

\newcommand{\ignore}[1]{}

\makeatletter
\renewcommand{\paragraph}{%
  \@startsection{paragraph}{4}%
  {\z@}{1.25ex \@plus 1ex \@minus .2ex}{-1em}%
  {\normalfont\normalsize\bfseries}%
}
\makeatother

\begin{document}
\maketitle
\begin{abstract}

We present an analysis tool based on joint matrix factorization for comparing latent representations of multilingual and monolingual models. An alternative to probing, this tool allows us to analyze multiple sets of representations in a joint manner. Using this tool, we study to what extent and how morphosyntactic features are reflected in the representations learned by multilingual pre-trained models. We conduct a large-scale empirical study of over 33 languages and 17 morphosyntactic categories. Our findings demonstrate variations in the encoding of morphosyntactic information across upper and lower layers, with category-specific differences influenced by language properties. Hierarchical clustering of the factorization outputs yields a tree structure that is related to phylogenetic trees manually crafted by linguists. Moreover, we find the factorization outputs exhibit strong associations with performance observed across different cross-lingual tasks. 
We release our code to facilitate future research.\footnote{\url{https://github.com/zsquaredz/joint_multilingual_analysis/}}

\end{abstract}

\section{Introduction}
Pre-trained multilingual models \citep{lample-2019-cross, conneau-etal-2020-unsupervised, liu-etal-2020-multilingual, xue-etal-2021-mt5} have gained widespread adoption in recent years. They initially pre-trained in many languages and subsequently fine-tuned for specific downstream tasks. Their aim is to leverage the linguistic knowledge acquired from similar languages, thereby benefiting low-resource languages and enabling zero-shot cross-lingual transfer ability. While numerous prior works have demonstrated these models have such abilities \citep{gerz-etal-2018-relation,ziser-reichart-2018-deep,aharoni-etal-2019-massively,karthikeyan-etal-2020-ability,muller-etal-2021-first,fujinuma-etal-2022-match,qiu2023detecting}, there are still open questions about the nature of the linguistic knowledge these models possess and the extent to which they acquire and incorporate linguistic information in their multilingual representations.

\begin{figure}[t]
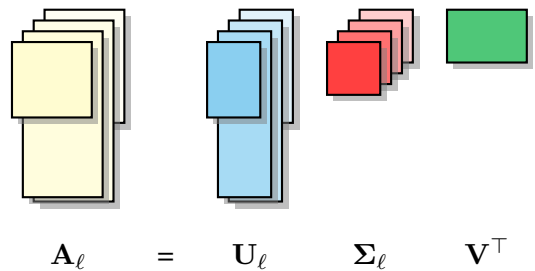

    \centering

\definecolor{babyblue}{rgb}{0.54, 0.81, 0.94}
\definecolor{cream}{rgb}{1.0, 0.99, 0.82}
\definecolor{coralred}{rgb}{1.0, 0.25, 0.25}
\definecolor{emerald}{rgb}{0.31, 0.78, 0.47}

\begin{tabular}{ccccc}
\rstack[1.05cm]{1.5,2.4,2.2,1}{cream}{} & & \rstack[0.7cm]{1.5,2.4,2.2,1}{babyblue}\qquad{} &\rstack[0.7cm]{0.7,0.7,0.7,0.7}{coralred}{} &\rstack[1.05cm]{0.7}{emerald}{} \\
\\
$\mymat{A}_\ell$ & = & $\mymat{U}_\ell$ & $\mymat{\Sigma}_\ell$ & $\mymat{V}^{\top}$ 
\end{tabular}

    \caption{A diagram of the matrix factorization that PARAFAC2 performs. For our analysis, $\mymat{A}_\ell$ ranges over covariance matrices between multilingual model representations and a $\ell$th monolingual model representations.}
    \label{fig:parafac2}
\end{figure}

Previous work has used singular vector canonical correlation analysis (SVCCA; \citealt{raghu-etal-2017-svcca}) and other similarity statistics like centered kernel alignment \citep[CKA;][]{kornblith-etal-2019-similarity} to analyze multilingual representations \citep{singh-etal-2019-bert,kudugunta-etal-2019-investigating,muller-etal-2021-first}. However, such methods can only compare one pair of representation sets at a time. In contrast to that, we analyze all multilingual representations simultaneously using parallel factor analysis 2 (PARAFAC2; \citealt{harshman-1972b-parafac2}), a method that allows us to factorize a set of representations jointly by decomposing it into multiple components that can be analyzed individually and then recombined to understand the underlying structure and patterns in the data. More precisely, we extend the sub-population analysis method recently presented by \citet{zhao-etal-2022-understanding}, who compare two models as an alternative to probing: a \emph{control model} trained on data of interest and an \emph{experimental model}, which is identical to the control model but is trained on additional data from different sources. By treating the multilingual experimental model as a shared component in multiple comparisons with different control models (each one is a monolingual model trained on a subset of the multilingual model), we can better analyze the multilingual representations.

As an alternative to probing, our representation analysis approach: a) enables standardized comparisons across languages within a multilingual model, circumventing the need for external performance upper bounds in making meaningful interpretation of performance metrics; b) directly analyzes model representations, avoiding the need for auxiliary probing models and potential biases from specific probing classifier architectures; and c) compares multilingual versus monolingual representations for any inputs, avoiding reliance on labelled probing datasets.

We use PARAFAC2 to directly compare representations learned between multilingual models and their monolingual counterparts. We apply this efficient paradigm to answer the following research questions on multilingual models: \textbf{Q1)} How do multilingual language models encode morphosyntactic features in their layers? \textbf{Q2)} Are our findings robust to address low-resource settings? \textbf{Q3)} Do morphosyntactic typology and downstream task performance reflect in the factorization outputs?

We experiment with two kinds of models, XLM-R \citep{conneau-etal-2020-unsupervised} and RoBERTa \cite{Liu2019RoBERTaAR}. We apply our analysis tool on the multilingual and monolingual representations from these models to check morphosyntactic information in 33 languages from Universal Dependencies treebanks \citep[UD;][]{nivre-2017-ud}. Our analysis reinforces recent findings on multilingual representations, such as the presence of language-neutral subspaces in multilingual language models \cite{foroutan-etal-2022-discovering}, and yields the following key insights:
\begin{itemizesquish}{-0.2em}{0.5em}
    \item Encoding of morphosyntactic information is influenced by language-specific factors such as writing system and number of unique characters.
    \item Multilingual representations demonstrate distinct encoding patterns in subsets of languages with low language proximity.
    \item Representation of low-resource languages benefits from the presence of related languages.
    \item Our factorization method's utility reflects in hierarchical clustering within phylogenetic trees and prediction of cross-lingual task performance.
\end{itemizesquish}

\section{Background and Motivation}
\label{section:background}

In this paper, we propose to use PARAFAC2 for multilingual analysis. By jointly decomposing a set of matrices representing cross-covariance between multilingual and monolingual representations, PARAFAC2 allows us to compare the representations across languages and their relationship to a multilingual model. For an integer $n$, we use $[n]$ to denote the set $\{1,...,n\}$. For a square matrix $\mymat{\Sigma}$, we denote by $\mathrm{diag}(\mymat{\Sigma})$ its diagonal vector.

\paragraph{PARAFAC2}
Let $\ell$ index a set of matrices,\footnote{In our case, $\ell$ will usually index a language. We will have several collections of such matrices indexed by language, for example, slicing the representations from a specific morphosyntactic category. This becomes clear in \S\ref{section:method} and \S\ref{section:experiments}.} such that $\mymat{A}_\ell = \mathbb{E}[\rv{X}_\ell\rv{Z}^{\top}]$, the matrix of cross-covariance between $\rv{X}_\ell$ and $\rv{Z}$, which are random vectors of dimensions $d$ and $d'$, respectively.\footnote{We assume that for any $\ell$, $\mathbb{E}[\rv{X}_\ell] = 0$ and $\mathbb{E}[\rv{Z}] = 0$.} This means that $[\mymat{A}_\ell]_{ij} = \mathrm{Cov}([\rv{X}_{\ell}]_i,\elementrv{Z}_j)$ for $i \in [d]$ and $j \in [d']$.

For any $\ell$ and two vectors, $\myvec{a} \in \mathbb{R}^d, \myvec{b} \in \mathbb{R}^{d'}$, the following holds due to the linearity of expectation:
$\myvec{a} \mymat{A}_\ell \myvec{b}^{\top} = \mathrm{Cov}(\myvec{a}^{\top} \rv{X}_\ell, \myvec{b}^{\top}\rv{Z}).$
PARAFAC2 on the set of matrices $\{ \mymat{A}_\ell \}_\ell$ in this case finds
a set of transformations $\{ \mymat{U}_\ell \}_\ell$, $\mymat{V}$ and a set of diagonal matrices $\{ \mymat{\Sigma}_\ell \}_\ell$ such that:

\begin{equation}
\mymat{A}_\ell \approx \mymat{U}_\ell \mymat{\Sigma}_\ell \mymat{V}^{\top}. \label{eq:BB}
\end{equation}

 We call the elements on the diagonal of $\mymat{\Sigma}_\ell$ \emph{pseudo-singular values}, in relationship to singular value decomposition that decomposes a single matrix in a similar manner.
The decomposition in Eq.~\refeq{eq:BB} jointly decomposes the matrices such that each $\mymat{A}_\ell$ is decomposed into a sequence of three transformations: first transforming $\rv{Z}$ into a latent space ($\mymat{V}$), scaling it ($\mymat{\Sigma}_\ell$) and then transforming it into a specific $\ell$th-indexed $\rv{X}_\ell$ space ($\mymat{U}_\ell$). Unlike singular value decomposition, which decomposes a matrix into a similar sequence of transformations with orthonormal matrices, PARAFAC2 does not guarantee $\mymat{U}_\ell$ and $\mymat{V}$ to be orthonormal and hence they do not represent an orthonormal basis transformation. However, \citet{harshman-1972b-parafac2} showed that a solution can still be found and is unique if we add the constraint that $\mymat{U}_\ell^{\top} \mymat{U}_\ell$ is constant for all $\ell$. In our use of PARAFAC2, we follow this variant. We provide an illustration of PARAFAC2 in Figure~\ref{fig:parafac2}.

\section{Experiment-Control Modeling for Multilingual Analysis}
\label{section:method}

We employ factor analysis to generate a distinctive signature for a group of representations within an experimental model, contrasting them with representations derived from a set of control models. In our case, the experimental model is a jointly-trained multilingual pre-trained language model, and the control models are monolingual models trained separately for a set of languages. Formally, there is an index set of languages $[L]$ and a set of models consisting of the experimental model $\mathbf{E}$ and the control models $\mathbf{C}_{\ell}$ for $\ell \in [L]$.

We assume a set of inputs we apply our analysis to, $\mathcal{X} = \bigcup_{\ell=1}^{L} \mathcal{X}_{\ell}$. Each set $\mathcal{X}_{\ell} = \{ \myvec{x}_{\ell,1}, \ldots, \myvec{x}_{\ell,m} \}$ represents a set of inputs for the $\ell$th language. While we assume, for simplicity, that all sets have the same size $m$, it does not have to be the case.
In our case, each $\mathcal{X}_{\ell}$ is a set of input words from language $\ell$, which results in a set of representations as follows. For each $\ell \in [L]$ and $i \in [m]$ we apply the model $\mathbf{E}$ and the model $\mathbf{C}_{\ell}$ to $\myvec{x}_{\ell,i}$
to get two corresponding representations $\myvec{y}_{\ell,i} \in \mathbb{R}^d$ and $\myvec{z}_{\ell,i} \in \mathbb{R}^{d_{\ell}}$. Here, $d$ is the dimension of the multilingual model and $d_{\ell}$ is the dimension of the representation of the monolingual model for the $\ell$th language.
Stacking up these sets of vectors separately into two matrices (per language $\ell$), we obtain the set of paired matrices $\mymat{Y}_{\ell} \in \mathbb{R}^{m \times d}$ and $\mymat{Z}_{\ell} \in \mathbb{R}^{m \times d_{\ell}}$. We further calculate the covariance matrix $\mymat{\Omega}_{\ell}$, {defined as:}
$\mymat{\Omega}_{\ell} = \mymat{Z}_{\ell}^{\top} \mymat{Y}_{\ell}$.

\paragraph{Use of PARAFAC2} Given an integer $k$ smaller or equal to the dimensions of the covariance matrices, we apply PARAFAC2 on the set of joint matrices, decomposing each $\mymat{\Omega}_{\ell}$ into:

\begin{equation}
\mymat{\Omega}_{\ell} \approx \mymat{U}_{\ell} \mymat{\Sigma}_{\ell} \mymat{V}^{\top} \text{,}\label{eq:AA}
\end{equation}

\noindent where $\mymat{U} \in \mathbb{R}^{d_{\ell} \times k}$ and $\mymat{V} \in \mathbb{R}^{d \times k}$.

To provide some intuition on this decomposition, consider Eq.~\refeq{eq:AA} for a fixed $\ell$.
If we were following SVD, such decomposition would give two projections that
project the multilingual representations and the monolingual representations into a joint latent space (by applying $\mymat{U}_{\ell}$ and $\mymat{V}$ on $\myvec{z}$s and $\myvec{y}$s, respectively). When applying PARAFAC2 jointly on the set of $L$ matrices, we enforce the matrix $\mymat{V}$ to be identical for all
decompositions (rather than be separately defined if we were applying SVD on each matrix separately) and for $\mymat{U}_{\ell}$ to vary for each language. We are now approximating the $\mymat{\Omega}_{\ell}$ matrix, which by itself could be thought as transforming vectors from the multilingual space to the monolingual space (and vice versa) in three transformation steps: first into a latent space ($\mymat{V}$), scaling it ($\mymat{\Sigma}_{\ell}$), and then \emph{specializing} it monolingually.

The diagonal of $\mymat{\Sigma}_{\ell}$ can now be readily used to describe a
\emph{signature} of the $\ell$th language representations in relation to the multilingual model (see also \citealt{dubossarsky-etal-2020-secret}). This signature, which we mark by $\mathrm{sig}(\ell) = \mathrm{diag}(\mymat{\Sigma}_{\ell})$,
can be used to compare the nature of representations between languages, and their commonalities in relationship to the multilingual model.
In our case, this PARAFAC2 analysis is applied to different slices of the data. We collect tokens in different languages (both through a multilingual model and monolingual models) and then slice them by specific morphosyntactic category, each time applying PARAFAC2 on a subset of them.

For some of our analysis, we also use a condensed value derived from $\mathrm{sig}(\ell)$. We follow a similar averaging approach to that used by SVCCA \citep{raghu-etal-2017-svcca}, a popular representation analysis tool, where they argue that the single condensed SVCCA score represents the average correlation across aligned directions and serves as a direct multidimensional analogue of Pearson correlation. In our case, each signature value within $\mathrm{sig}(\ell)$ from the PARAFAC2 algorithm corresponds to a direction, all of which are normalized in length, so the signature values reflect their relative strength. Thus, taking the average of $\mathrm{sig}(\ell)$ provides an intensity measure of the representation of a given language in the multilingual model. We provide additional discussion in \S\ref{section:morphrsyntatic_language_property}.

\section{Experimental Setup}
\label{section:experiments}

\paragraph{Data} We use CoNLL's 2017 Wikipedia dump \citep{ginter-etal-2017-conll} to train our models. Following \citet{fujinuma-etal-2022-match}, we downsample all Wikipedia datasets to an identical number of sequences to use the same amount of pre-training data for each language.  In total, we experiment with 33 languages. For morphosyntactic features, we use treebanks from UD 2.1 \citep{nivre-2017-ud}. These treebanks contain sentences annotated with morphosyntactic information and are available for a wide range of languages. We obtain a representation for every word in the treebanks using our pre-trained models. We provide further details on our pre-training data and how we process morphosyntactic features in Appendix~\ref{app:exp_data}.

\paragraph{Task} For pre-training our models, we use masked language modeling (MLM). Following \citet{devlin-etal-2019-bert}, we mask 15\% of the tokens. 
To fully control our experiments, we follow \citet{zhao-etal-2022-understanding} and train our models from scratch. 

\paragraph{Models} We have two kinds of models: the multilingual model $\mymat{E}$, trained using all $L$ languages available, and the monolingual model $\mymat{C}_\ell$ for $\ell \in [L]$ trained only using the $\ell$th language. We use the XLM-R \citep{conneau-etal-2020-unsupervised} architecture for the multilingual $\mymat{E}$ model, and we use RoBERTa \citep{Liu2019RoBERTaAR} for the monolingual $\mymat{C}_{\ell}$ model. We use the base variant for both kinds of models. We use XLM-R's vocabulary and the SentencePiece \citep{kudo-richardson-2018-sentencepiece} tokenizer for all our experiments provided by \citet{conneau-etal-2020-unsupervised}. This enables us to support all languages we analyze and ensure fair comparison for all configurations. We provide additional details about our models and training in Appendix~\ref{app:exp_model}.

\section{Experiments and Results}
This section outlines our research questions (RQs), experimental design, and obtained results.

\subsection{Morphosyntactic and Language Properties}
\label{section:morphrsyntatic_language_property}
Here, we address RQ1: \textit{How do multilingual language models encode morphosyntactic features in their layers?} While in broad strokes, previous work \citep{hewitt-manning-2019-structural,jawahar-etal-2019-bert,tenney-etal-2019-bert} showed that syntactic information tends to be captured in lower to middle layers within a network, we ask a more refined question here, and inspect whether different layers are specialized for specific morphosyntactic features, rather than providing an overall picture of all morphosyntax in a single layer. As mentioned in \S\ref{section:method}, we have a set of signatures, $\mathrm{sig}(\ell)$ for $\ell \in [L]$, each describing the $\ell$th language representation for the corresponding morphosyntactic category we probe and the extent to which it utilizes information from each direction within the rows of $\mymat{V}$.
PARAFAC2 identifies a single transformation $\mymat{V}$ that maps a multilingual representation into a latent space. Following that, the signature vector scales in specific directions based on their importance for the final monolingual representation it is transformed to. Therefore, the signature can be used to analyze whether similar directions in $\mymat{V}$ are important for the transformation to the monolingual space.
By using signatures of different layers in a joint factorization, we can identify comparable similarities for all languages. Analogous to the SVCCA similarity score \citep{raghu-etal-2017-svcca}, we condense each signature vector into a single value by taking the average of the signature. This value encapsulates the intensity of the use of directions in $\mymat{V}$. A high average indicates the corresponding language is well-represented in the multilingual model. 
We expect these values to exhibit a general trend (either decreasing or increasing) going from lowers to upper layers as lower layers are more general and upper layers are known to be more task-specific \citep{rogers-etal-2020-primer}. In addition, the trend may be contrasting for different languages and morphosyntactic features. 

\begin{figure}[t]
     \centering
     \begin{subfigure}[b]{0.23\textwidth}
         \centering
         \includegraphics[width=\textwidth]{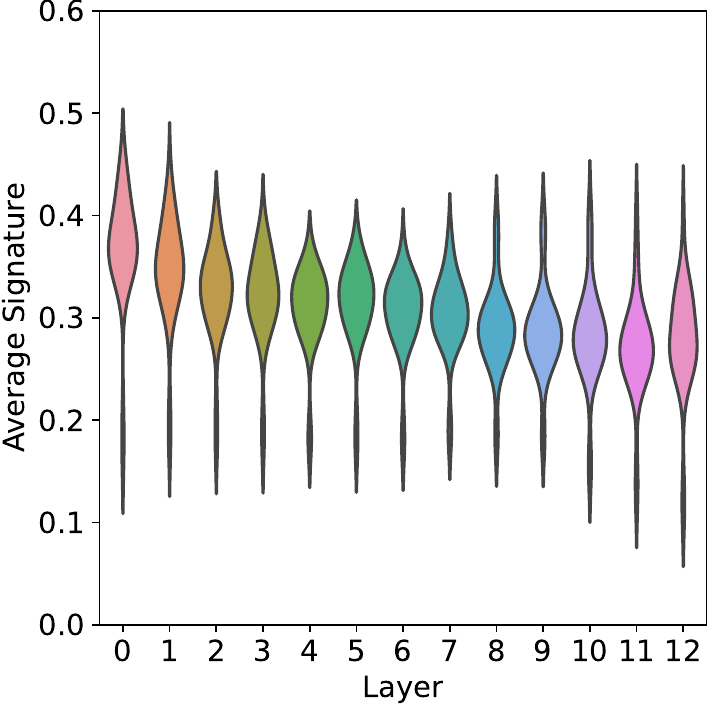}
         \caption{Original data}
         \label{fig:pos_violin_no_trans}
     \end{subfigure}
    \hfill
     \begin{subfigure}[b]{0.23\textwidth}
         \centering
         \includegraphics[width=\textwidth]{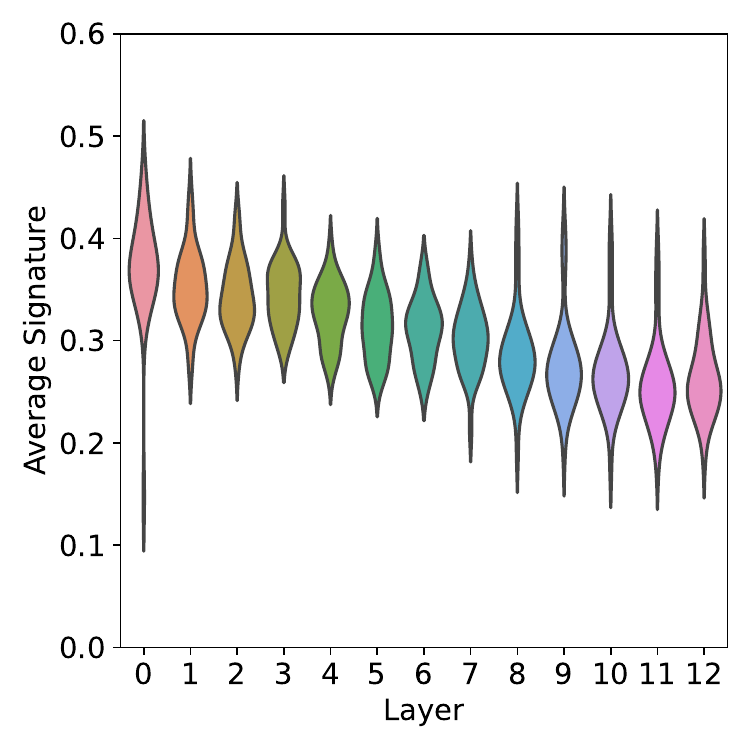}
         \caption{\texttt{zh} and \texttt{ja} romanized data}
         \label{fig:pos_violin_trans}
     \end{subfigure}
\caption{Average signature violin plots for all layers and languages on (a) original data and (b) data with Chinese (\texttt{zh}) and Japanese (\texttt{ja}) romanized.}
\label{fig:pos_violin}
\end{figure}

\paragraph{Language Signatures Across Layers}
We begin by presenting the distribution of average $\mathrm{sig}(\ell)$ values for all languages across all layers for all lexical tokens in Figure~\ref{fig:pos_violin_no_trans}. 
We observe a gradual decrease in the mean of the distribution as we transition from lower to upper layers.
This finding is consistent with those from \citet{singh-etal-2019-bert}, who found that the similarity between representations of different languages steadily decreases up to the final layer in a pre-trained mBERT model.
We used the Mann-Kendall (MK) statistical test \citep{mann1945nonparametric,kendall1948rank} for individual languages across all layers. The MK test is a rank-based non-parametric method used to assess whether a set of data values is increasing or decreasing over time, with the null hypothesis being there is no clear trend. Since we perform multiple tests (33 tests in total for all languages), we also control the false discovery rate (FDR; at level $q = 0.05$) with corrections to the $p$-values \citep{benjamini1995controlling}. We found that all 33 languages except for Arabic, Indonesian, Japanese, Korean, and Swedish exhibit significant monotonically decreasing trends from lower layers to upper layers, with the FDR-adjusted $p$-values ($p< 0.05$). 
Figure~\ref{fig:pos_violin_no_trans} shows that the spread of the distribution for each layer (measured in variance) is constantly decreasing up until layer 6. From these layers forward, the spread increases again. A small spread indicates that the average intensity of scaling from a multilingual representation to the monolingual representation is similar among all languages. This provides evidence of the multilingual model aligning languages into a \emph{language-neutral} subspace in the middle layers, with the upper layers becoming more task-focused \citep{merchant-etal-2020-happens}. This result is also supported by findings of \citet{muller-etal-2021-first} -- different languages representations' similarity in mBERT constantly increases up to a mid-layer then decreases.

\paragraph{Logogram vs. Phonogram}
In Figure~\ref{fig:pos_violin_no_trans}, we observe a long bottom tail in the average $\mathrm{sig}(\ell)$ plots for all languages, with Chinese and Japanese showing lower values compared to other languages that are clustered together, suggesting that our models have learned distinct representations for those two languages. 
We investigated if this relates to the logographic writing systems of these languages, which rely on symbols to represent words or morphemes rather than phonetic elements. We conducted an ablation study where we romanized our Chinese and Japanese data into Pinyin and Romaji,\footnote{We use libraries available at: \url{https://pypi.org/project/pypinyin/} and \url{https://pypi.org/project/pykakasi/}. We use the \texttt{lazy\_pinyin} feature to generate Pinyins without tone marks.} respectively, and retrained our models. 
One might ask why we did not normalize the other languages in our experiment to use the Latin alphabet. There are two reasons for this: 1) the multilingual model appears to learn them well, as evidenced by their similar signature values to other languages; 2) our primary focus is on investigating the impact of logographic writing systems, with Chinese and Japanese being the only languages employing logograms, while the others use phonograms.
Figure~\ref{fig:pos_violin_trans} shows that, apart from the embedding layer, the average $\mathrm{sig}(\ell)$ are more closely clustered together after the ablation. 
Our findings suggest that logographic writing systems may present unique challenges for multilingual models, warranting further research to understand their computational processes.
Although not further explored here, writing systems should be considered when developing and analyzing multilingual models.

\begin{figure}[t]
     \centering
    \includegraphics[width=0.9\linewidth]{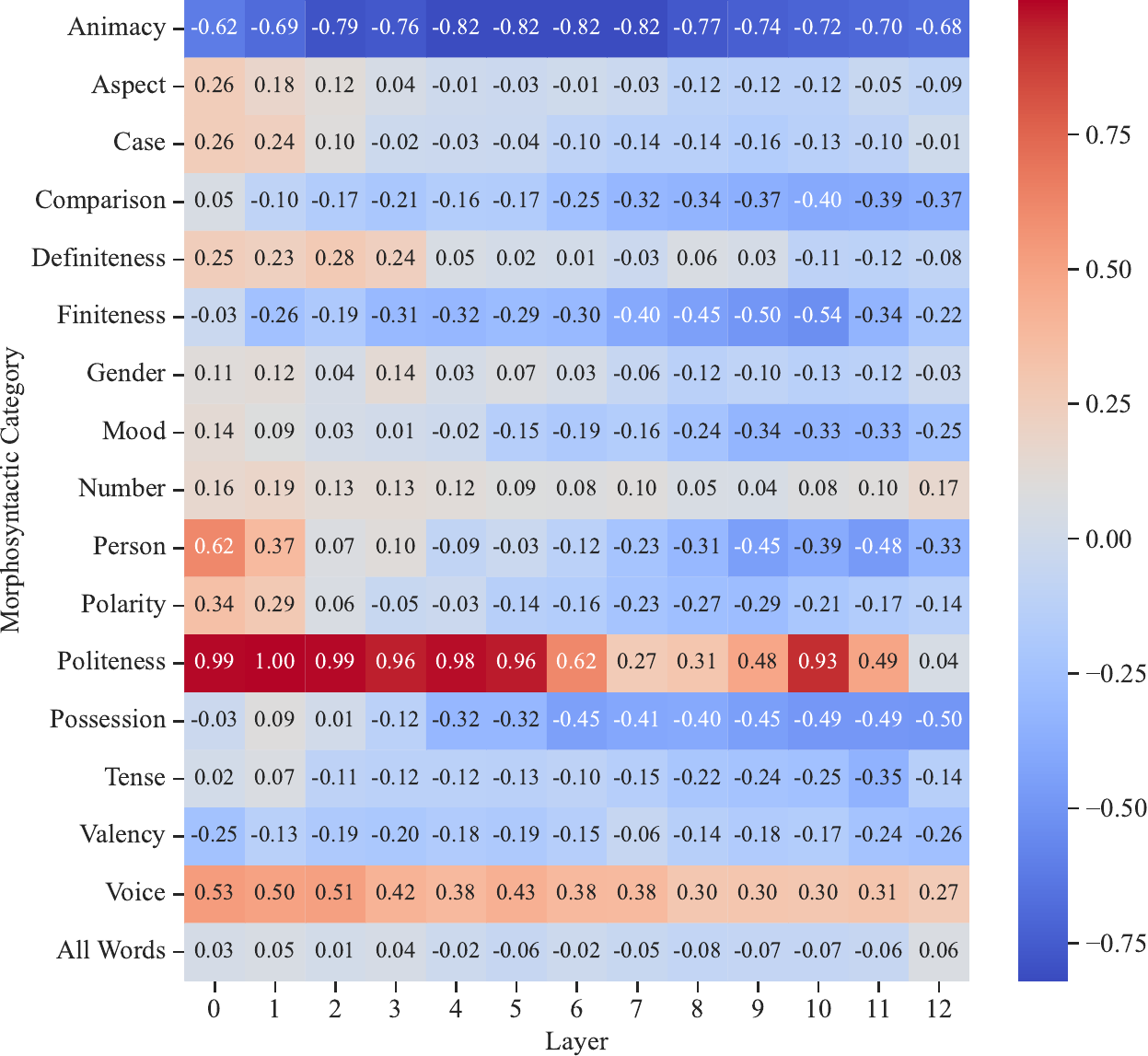}       
\caption{Pearson correlation results between the average $\mathrm{sig}(\ell)$ for all languages and their data size for each morphosyntactic category among all layers.}
\label{fig:correlation_data_size_all_layers}
\end{figure}

\paragraph{Morphosyntactic Attributes}
Looking at individual morphosyntactic attributes, we observe that while most attributes exhibit a similar decreasing trend from lower to upper layers, some attributes, such as Comparison and Polarity, show consistent distributions across all layers. 
Since these attributes occur rarely in our data ($<$ 1\% of tokens), it is possible that the model is only able to learn a general representation and not distinguish them among the layers.
To investigate the effect of attribute frequency on our analysis, we performed a Pearson correlation analysis (for each attribute) between the average $\mathrm{sig}(\ell)$ for all languages and their data size -- the number of data points available in the UD annotations for a particular language and morphosyntactic feature. 
The results are shown in Figure~\ref{fig:correlation_data_size_all_layers}. 
Our analysis of the overall dataset (all words) shows no evidence of correlation between attribute frequency and average $\mathrm{sig}(\ell)$.
However, upon examining individual categories, we observe a decrease in correlation as we move up the layers, indicating that the degree a morphosyntactic attribute is represented in the multilingual model is no longer associated with simple features like frequency but rather with some language-specific properties. 
This observation holds true for all categories, with the exception of Animacy, which is predominantly found in Slavic languages within our dataset. This aligns with the findings of \citet{stanczak-etal-2022-neurons}, who noted that the correlation analysis results can be affected by whether a category is typical for a specific genus. Next, we further explore the relationship between signature values and language properties. 

\begin{figure*}[ht]
    \centering
    \includegraphics[width=\linewidth]{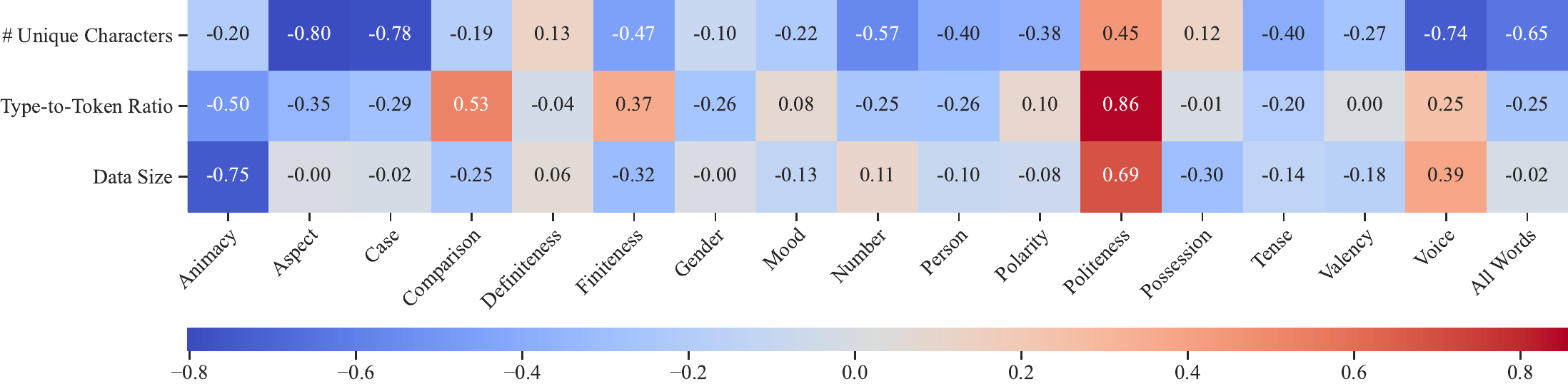}       
\caption{Pearson correlation results between the average $\mathrm{sig}(\ell)$ for all languages and the number of unique characters, type-token ratio (TTR), and data size for each morphosyntactic category, averaged across all layers.}
\label{fig:correlation_lang_property}
\end{figure*}

\paragraph{Language Properties}
In addition to data size, we explore the potential relationship between language-specific properties and the average $\mathrm{sig}(\ell)$. We consider two language properties: the number of unique characters and the type-token ratio (TTR), a commonly used linguistic metric to assess a text's vocabulary diversity.\footnote{To ensure accurate analysis, we filter out noise by counting the number of characters that account for 99.9\% of occurrences in the training data. This eliminates characters that only appear very few times.}  TTR is calculated by dividing the number of unique words (measured in lemmas) by the total number of words (measured in tokens) obtained from the UD annotation meta-data. Typically, a higher TTR indicates a greater degree of lexical variation. We present the Pearson correlation, averaged across all layers, in Figure~\ref{fig:correlation_lang_property}. To provide a comprehensive comparison, we include the results for data size as well. The detailed results for each layer can be found in Appendix~\ref{app:rq1}. Examining the overall dataset, we observe a strong negative correlation between the number of unique characters and signature values. Similarly, the TTR exhibits a similar negative correlation, indicating that higher language variation corresponds to lower signature values. When analyzing individual categories, we consistently find a negative correlation for both the number of unique characters and the TTR. This further supports our earlier finding that Chinese and Japanese have lower signature values compared to other languages, as they possess a higher number of unique characters and TTR.

\paragraph{Generalization to Fully Pre-trained Models}
To ensure equal data representation for all languages in our experiment-controlled modeling, we downsampled the Wikipedia dataset and used an equal amount for pre-training our multilingual models. To check whether our findings are also valid for multilingual pre-trained models trained on full-scale data, we conducted additional experiments using a public XLM-R checkpoint.\footnote{\url{https://huggingface.co/xlm-roberta-base}} The setup remained the same, except that we used representations obtained from this public XLM-R instead of our own trained XLM-R. We observe that the trends for signature values were generally similar, except for the embedding and final layers, where the values were very low. This was expected, as the cross-covariance was calculated with our monolingual models. The similar trend among the middle layers further supports the idea that these layers learn language- and data-agnostic representations. Furthermore, the Pearson correlations between the number of unique characters, TTR, data size, and the average $\mathrm{sig}(\ell)$ for the overall dataset were as follows: -0.65, -0.28, and -0.02, respectively. These values are nearly identical to those shown in Figure~\ref{fig:correlation_lang_property}, confirming the robustness of our method and its data-agnostic nature.

\begin{figure*}[ht]
     \centering
     \begin{subfigure}[b]{0.3\textwidth}
         \centering
         \includegraphics[width=\textwidth]{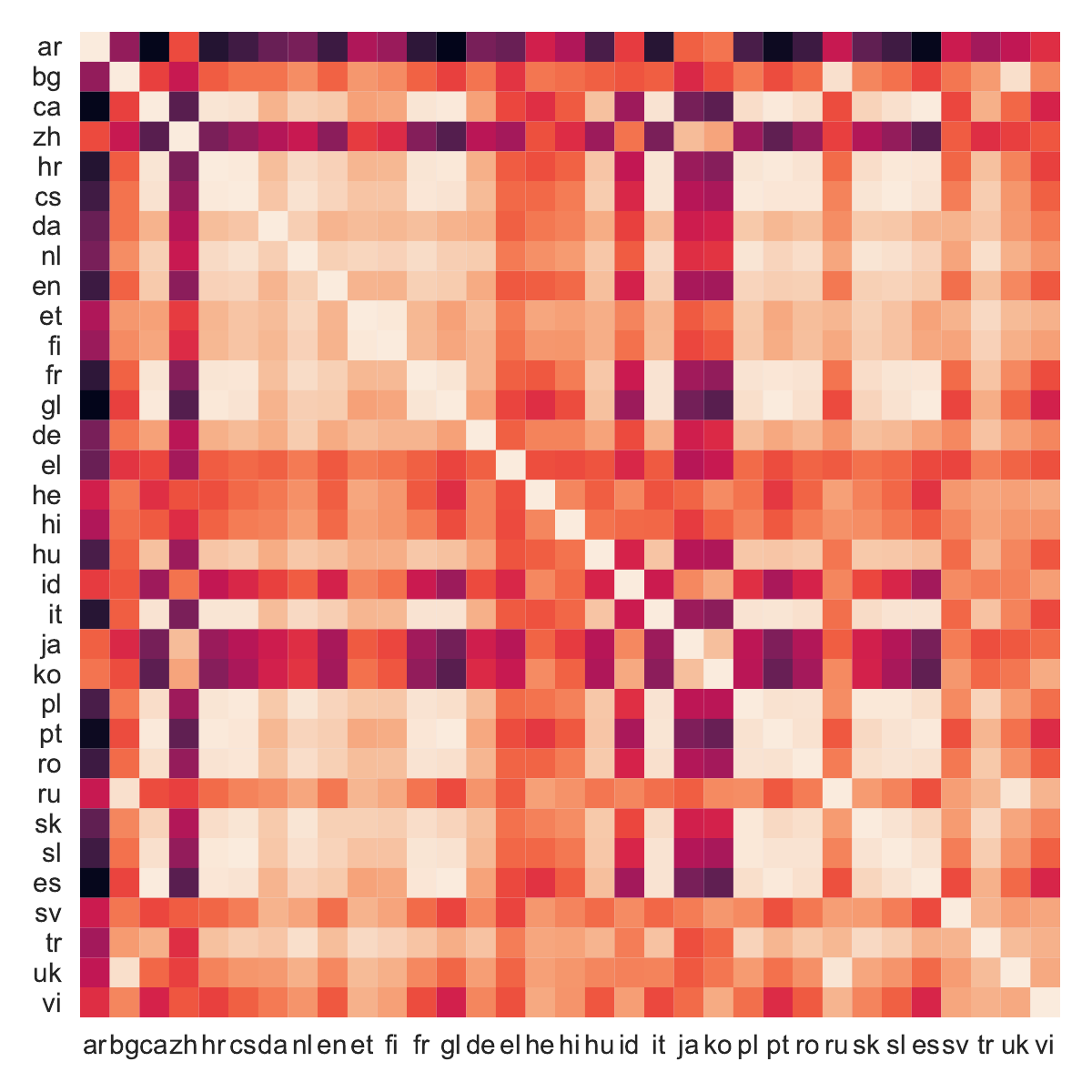}
         \caption{Layer 0}
         \label{fig:dist_matrix_l0}
     \end{subfigure}
    \hfill
     \begin{subfigure}[b]{0.3\textwidth}
         \centering
         \includegraphics[width=\textwidth]{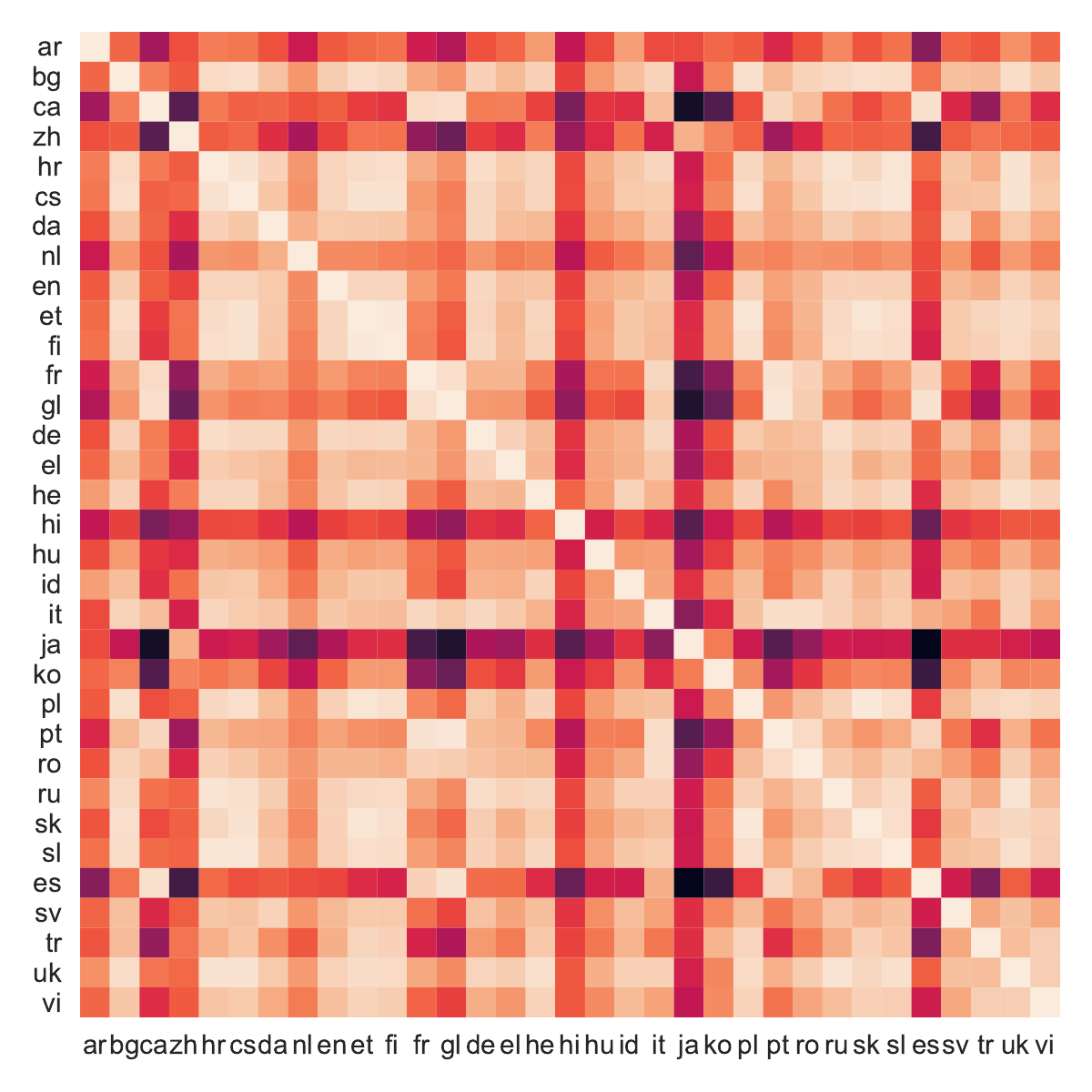}
         \caption{Layer 6}
         \label{fig:dist_matrix_l6}
     \end{subfigure}
    \hfill
     \begin{subfigure}[b]{0.3\textwidth}
         \centering
         \includegraphics[width=\textwidth]{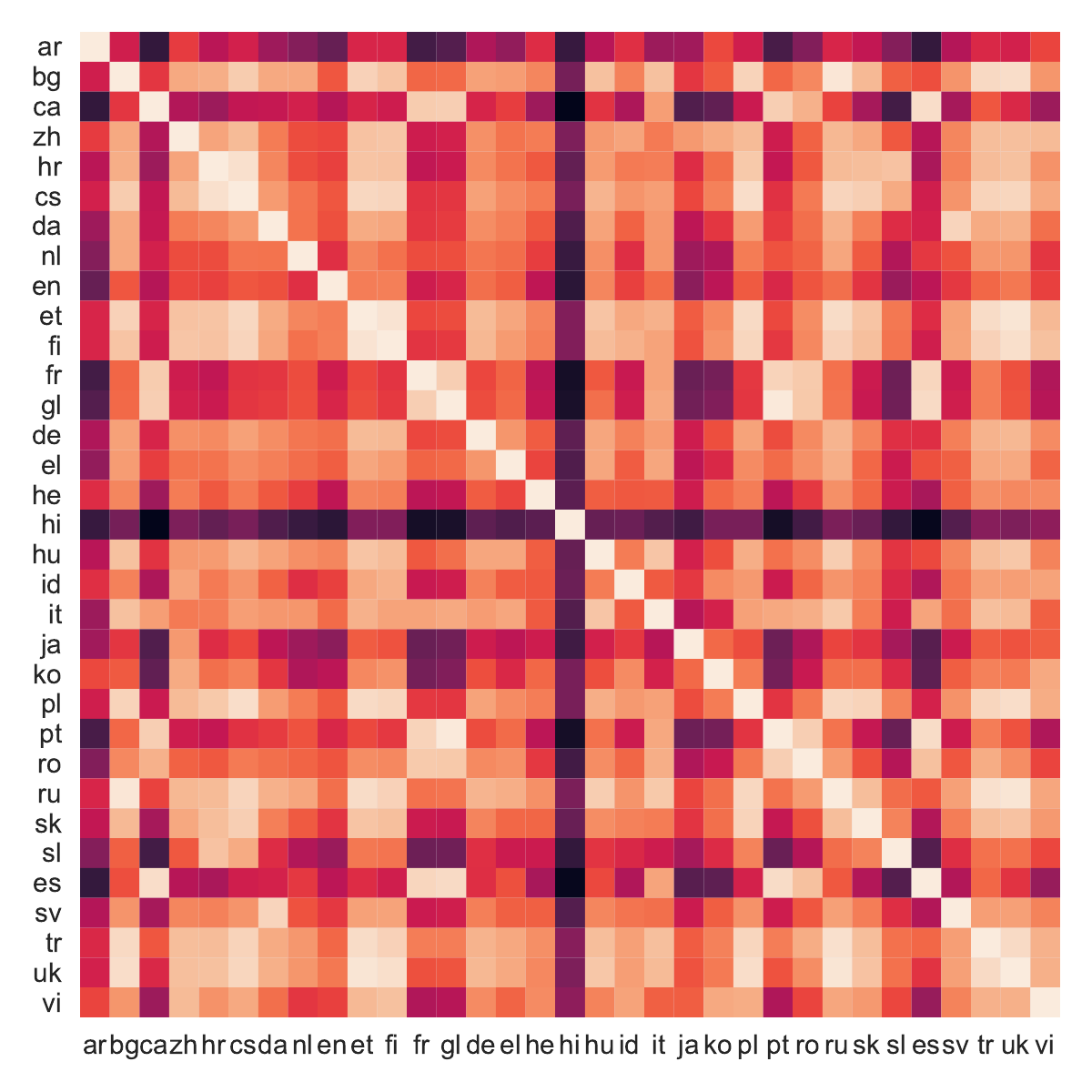}
         \caption{Layer 12}
         \label{fig:dist_matrix_l12}
     \end{subfigure}
\caption{Cosine distance matrices between all language pairs and their signature vectors based on overall representations obtained from layer 0, 6 and 12. Darker color indicates the cosine distance being close to 1.}
\label{fig:dist_matrix_all}
\end{figure*}

\subsection{Language Proximity and Low-resource Conditions}

Here, we address RQ2: \textit{Are our findings robust to address language subsets and low-resource settings?} In RQ1, our analysis was based on the full set of pre-training languages available for each morphosyntactic category we examine. In this question, we aim to explore subsets of representations derived from either a related or diverse set of pre-training languages, and whether such choices yield any alterations to the findings established in RQ1. Furthermore, we extend our analysis to low-resource settings and explore potential changes in results on low-resource languages, particularly when these languages could receive support from other languages within the same language family. We also explore the potential benefits of employing language sampling techniques for enhancing the representation of low-resource languages.

\paragraph{Language Proximity} 
We obtain the related set of languages by adding all languages that are from the same linguistic family and genus (full information available in Appendix~\ref{app:appendix}). In total, we obtained three related sets of languages: Germanic languages, Romance languages, and Slavic languages. There are other related sets, but we do not include them in our experiment since the size of those sets is very small. For the diverse set of languages, we follow \citet{fujinuma-etal-2022-match} and choose ten languages from different language genera that have a diverse set of scripts . These languages are Arabic, Chinese, English, Finnish, Greek, Hindi, Indonesian, Russian, Spanish, and Turkish. We use the $\chi^2$-square variance test to check whether the variance of the diverse set's average signatures from a particular layer is statistically significant from the variance of that of the related set, given a morphosyntactic category. We test layers 0 (the embedding layer), 6, and 12, covering the lower, middle, and upper layers within the model. We first find that for the overall dataset, the variance of the diverse set average signatures is significantly different (at $\alpha=0.05$) from all three related set variances for all three layers. This suggests that, in general, multilingual representations are encoded differently for different subsets of languages with low language proximity. 
For the attributes of number, person, and tense, the variance within the diverse set significantly differs from the variances within the three related sets across all three layers, with a statistical significance level of $\alpha=0.05$. This finding is sensible as all these three attributes have distinctions in the diverse set of languages. For example, Arabic has dual nouns to denote the special case of two persons, animals, or things, and Russian has a special plural form of nouns if they occur after numerals. On the other hand, for attributes like gender, we do not witness a significant difference between the diverse set and related set since there are only four possible values (\texttt{masculine}, \texttt{feminine}, \texttt{neuter}, and \texttt{common}) in the UD annotation for gender. We speculate that this low number of values leads to low variation among languages, thus the non-significant difference. This finding concurs with \citet{stanczak-etal-2022-neurons}, who observed a negative correlation between the number of values per morphosyntactic category and the proportion of language pairs with significant neuron overlap. Hence, the lack of significant differences in variance between the diverse and related sets can be attributed to the substantial overlap of neurons across language pairs.

\begin{table*}[t]
\centering
\small
\begin{tabular}{llrlcccc}
\Xhline{1pt}
Task                            & Dataset      &  \#Lang. & Metric   & mBERT     & XLM       & XLM-R     & MMTE      \\
\Xhline{1pt}
\multirow{2}{*}{Classification} & XNLI \cite{conneau-etal-2018-xnli}    & 12     & Acc.       & .36       & .30       & .36       & .21       \\
                                & PAWS-X \cite{yang-etal-2019-paws}   & 7   & Acc.       & .67       & .65       & .75       & .69       \\
                                \hline
\multirow{2}{*}{Struct. pred.}  & POS \cite{nivre-etal-udv2.2-2018}    & 22     & F1          & .36       & .36       & .66       & .40       \\
                                & NER \cite{pan-etal-2017-cross}      & 22   & F1        & .46       & .46       & .55       & .46       \\
                                \hline
\multirow{3}{*}{QA}             & XQuAD \cite{artetxe-etal-2020-cross}  & 10      & F1 / EM    & .60 / .35 & .81 / .56 & .73 / .45 & .72 / .61 \\
                                & MLQA \cite{lewis-etal-2020-mlqa}     & 7   & F1 / EM     & .23 / .31 & .46 / .48 & .64 / .68 & .28 / -   \\
                                & TyDiQA-GoldP \cite{clark-etal-2020-tydi} & 6 & F1 / EM     & .41 / .05 & .43 / .43 & .46 / .46 & .66 / .45 \\
                                \hline
\multirow{2}{*}{Retrieval}      & BUCC \cite{zweigenbaum-etal-2017-overview}  & 4      & F1      & .72           & .96       & .83       & .63       \\
                                & Tatoeba \cite{artetxe-schwenk-2019-massively} & 21    & Acc.    & .15         & .24       & .28       & -         \\

\Xhline{1pt}
\end{tabular}
\caption{Pearson correlations between final layer's $\mathrm{sig}(\ell)$ and XTREME benchmark performances on various tasks.}
\label{tab:correlation_xtreme}
\end{table*}

\paragraph{Low-resource Scenario}
In order to simulate a low-resource scenario, we curtailed the training data for selected languages, reducing it to only 10\% of its original size. The choice of low-resource languages included English, French, Korean, Turkish, and Vietnamese. English and French were selected due to the availability of other languages within the same language family, while the remaining languages were chosen for their absence of such familial relationships. Notably, Korean was specifically selected as it utilizes a distinct script known as Hangul. To examine the impact of low-resource conditions on each of the selected languages, we re-trained our multilingual model, with each individual language designated as low-resource. To address potential confounding factors, we also re-trained monolingual models on the reduced dataset. Additionally, we explored a sampling technique \cite{Devlin_2019} to enhance low-resource languages. Further details can be found in Appendix~\ref{app:rq2}.

Our analysis reveals the impact of low-resource conditions on signature values. English and French, benefiting from languages within the same language family, exhibit minimal changes in signature values, indicating a mitigation of the effects of low-resource conditions on language representation. Remaining languages without such support experience a significant decline in signature values (dropping from 0.3 to nearly 0), particularly after the embedding layer. This implies that low-resource languages struggle to maintain robust representations without assistance from related languages. Additionally, our findings suggest that language sampling techniques offer limited improvement in signature values of low-resource languages.

\subsection{Utility of Our Method}

Here we address RQ3: \textit{Do morphosyntactic typology and downstream task performance reflect in the factorization outputs?} Having conducted quantitative analyses of our proposed analysis tool thus far, our focus now shifts to exploring the tool's ability to unveil morphosyntactic information within multilingual representations and establish a relationship between the factorization outputs and downstream task performance. To investigate these aspects, we conduct two additional experiments utilizing the signature vectors obtained from our analysis tool. Firstly, we construct a phylogenetic tree using cosine distance matrices of all signature vectors. Subsequently, we examine the correlations between the results of the XTREME benchmark \cite{hu2020xtreme} and the $\mathrm{sig}(\ell)$ values.

\paragraph{Phylogenetic Tree} We first compute cosine distance matrices using all signature vectors for all 33 languages and 12 layers for each morphosyntactic attribute. Then, from the distance matrix, we use an agglomerative (bottom-up) hierarchical clustering method: unweighted pair group method with arithmetic mean \citep[UPGMA;][]{Sokal-1958-ASM} to construct a phylogenetic tree. We show the distance matrices between all language pairs and their signature vectors based on overall representations obtained from layers 0, 6 and 12 in Figure~\ref{fig:dist_matrix_all}. We can observe that signatures for Arabic, Chinese, Hindi, Japanese, and Korean are always far with respect to those for other languages across layers. From the distance matrix, we construct a phylogenetic tree using the UPGMA cluster algorithm. We present our generated trees and a discussion in Appendix~\ref{app:rq3-tree}. In short, the constructed phylogenetic tree resembles linguistically-crafted trees.

\paragraph{Performance Prediction} 
To establish a robust connection between our factorization outputs and downstream task performances, we conducted an analysis using the XTREME benchmark, which includes several models: mBERT \cite{devlin-etal-2019-bert}, XLM \cite{NEURIPS2019_xlm}, XLM-R, and MMTE \cite{arivazhgan-etal-2019-massively}. This benchmark encompasses nine tasks that span four different categories: classification, structured prediction, question answering, and retrieval. These tasks demand reasoning on multiple levels of meaning. To evaluate the relationship between the metrics of each task and our average $\mathrm{sig}(\ell)$ across all available languages for that task, we calculated the Pearson correlation. For each task's performance metrics, we use the results reported by \citet{hu2020xtreme}. The obtained correlation values using signature values from the last layer are presented in Table~\ref{tab:correlation_xtreme}, along with pertinent details about each task, such as the number of available languages, and the metrics employed. For a comprehensive analysis, we also provide results using $\mathrm{sig}(\ell)$ from every layer in Appendix~\ref{app:rq3-xtreme}. Observing the results, it becomes evident that the XLM-R model exhibits the highest correlation, which is expected since the $\mathrm{sig}(\ell)$ values obtained from our factorization process are also computed using the same architecture. Furthermore, for most tasks, the highest correlation is observed with the final layers, which is reasonable considering their proximity to the output. Notably, we consistently observe high correlation across all layers for straightforward tasks like POS and PAWS-X operating on the representation level. However, for complex reasoning tasks like XNLI, only the final layer achieves reasonable correlation. These results suggest that the factorization outputs can serve as a valuable indicator of performance for downstream tasks, even without the need for fine-tuning or the availability of task-specific data.

\section{Related Work}
Understanding the information within NLP models' internal representations has drawn increasing attention in the community. \citet{singh-etal-2019-bert} applied canonical correlation analysis (CCA) on the internal representations of a pre-trained mBERT and revealed that the model partitions representations for each language rather than using a shared interlingual space. 
\citet{kudugunta-etal-2019-investigating} used SVCCA to investigate massively multilingual Neural Machine Translation (NMT) representations and found that different language encoder representations group together based on linguistic similarity. 
\citet{Libovick-2019-HowLI} showed that mBERT representations could be split into a language-specific component and a language-neutral component by centering mBERT representations and using the centered representation on several probing tasks to evaluate the language neutrality of the representations. Similarly, \citet{foroutan-etal-2022-discovering} employed the lottery ticket hypothesis to discover sub-networks within mBERT and found that mBERT is comprised of language-neutral and language-specific components, with the former having a greater impact on cross-lingual transfer performance. \citet{muller-etal-2021-first} presented a novel layer ablation approach and demonstrated that mBERT could be viewed as the stacking of two sub-networks: a multilingual encoder followed by a task-specific language-agnostic predictor.

Probing (see \citealt{belinkov2022probing} for a review) is a widely-used method for analyzing multilingual representations and quantifying the information encoded by training a parameterized model, but its effectiveness can be influenced by model parameters and evaluation metrics \citep{pimentel-etal-2020-information}. \citet{choenni2020what} probed representations from multilingual sentence encoders and discovered that typological properties are
persistently encoded across layers in mBERT and XLM-R. \citet{liang2021locating} demonstrated with probing that language-specific information is scattered across many dimensions, which can be projected into a linear subspace. Intrinsic probing, on the other hand, explores the internal structure of linguistic information within representations \citep{torroba-hennigen-etal-2020-intrinsic}. \citet{stanczak-etal-2022-neurons} conducted a large-scale empirical study over two multilingual pre-trained models, mBERT, and XLM-R, and investigated whether morphosyntactic information is encoded in the same subset of neurons in different languages. 
Their findings reveal that there is considerable cross-lingual overlap between neurons, but the magnitude varies among categories and is dependent on language proximity and pre-training data size.
Other methods, such as matrix factorization techniques, are available for analyzing representations \citep{raghu-etal-2017-svcca,morcos-etal-2018-insights,kornblith-etal-2019-similarity} and even modifying them through model editing \citep{olfat2019convex,10.1162/tacl_a_00558,pmlr-v206-kleindessner23a,shao2022gold}. When applied to multilingual analysis, these methods are limited to pairwise language comparisons, whereas our proposed method enables joint factorization of multiple representations, making it well-suited for multilingual analysis.

\section{Conclusions}
We introduce a representation analysis tool based on joint matrix factorization.
We conduct a large-scale empirical study over 33 languages and 17 morphosyntactic categories and apply our tool to compare the latent representations learned by multilingual and monolingual models from the study. Our findings show variations in the encoding of morphosyntactic information across different layers of multilingual models. Language-specific differences contribute to these variations, influenced by factors such as writing systems and linguistic relatedness. 
Furthermore, the factorization outputs exhibit strong correlations with cross-lingual task performance and produce a phylogenetic tree structure resembling those constructed by linguists. These findings contribute to our understanding of language representation in multilingual models and have practical implications for improving performance in cross-lingual tasks. In future work, we would like to extend our analysis tool to examine representations learned by multimodal models.

\section*{Limitations}
Our research has several limitations. First, we only used RoBERTa and its multilingual variant, XLM-R, for our experiments. While these models are widely used in NLP research, there are other options available such as BERT, mBERT, T5, and mT5, which we have yet to explore due to a limited budget of computational resources. Second, to ensure equal data representation for all languages we experimented with, we downsampled Wikipedia resulting in a corpus of around 200MB per language. While we validated our findings against a publicly available XLM-R checkpoint trained on a much larger resource, further verification is still necessary. Third, our analyses are limited to morphosyntactic features, and in the future, we aim to expand our scope to include other linguistic aspects, such as semantics and pragmatics.

\section*{Acknowledgments}
This work was supported by the UKRI Centre for Doctoral Training (CDT) in Natural Language Processing through UKRI grant EP/S022481/1 and CDT funding from Huawei Technologies. We would like to thank Balint Gyevnar, Matthias Lindemann, Edoardo Ponti, Ivan Titov and the anonymous reviewers for their helpful feedback. We appreciate the use of computing resources through the CSD3 cluster at the University of Cambridge and the Baskerville cluster at the University of Birmingham.

\bibliography{anthology,custom}

\begin{thebibliography}{64}
\expandafter\ifx\csname natexlab\endcsname\relax\def\natexlab#1{#1}\fi

\bibitem[{Aharoni et~al.(2019)Aharoni, Johnson, and Firat}]{aharoni-etal-2019-massively}
Roee Aharoni, Melvin Johnson, and Orhan Firat. 2019.
\newblock \href {https://doi.org/10.18653/v1/N19-1388} {Massively multilingual neural machine translation}.
\newblock In \emph{Proceedings of the 2019 Conference of the North {A}merican Chapter of the Association for Computational Linguistics: Human Language Technologies, Volume 1 (Long and Short Papers)}, pages 3874--3884, Minneapolis, Minnesota. Association for Computational Linguistics.

\bibitem[{Arivazhagan et~al.(2019)Arivazhagan, Bapna, Firat, Lepikhin, Johnson, Krikun, Chen, Cao, Foster, Cherry, Macherey, Chen, and Wu}]{arivazhgan-etal-2019-massively}
Naveen Arivazhagan, Ankur Bapna, Orhan Firat, Dmitry Lepikhin, Melvin Johnson, Maxim Krikun, Mia~Xu Chen, Yuan Cao, George~F. Foster, Colin Cherry, Wolfgang Macherey, Zhifeng Chen, and Yonghui Wu. 2019.
\newblock \href {https://arxiv.org/abs/1907.05019} {Massively multilingual neural machine translation in the wild: Findings and challenges}.
\newblock \emph{ArXiv preprint}, abs/1907.05019.

\bibitem[{Artetxe et~al.(2020)Artetxe, Ruder, and Yogatama}]{artetxe-etal-2020-cross}
Mikel Artetxe, Sebastian Ruder, and Dani Yogatama. 2020.
\newblock \href {https://doi.org/10.18653/v1/2020.acl-main.421} {On the cross-lingual transferability of monolingual representations}.
\newblock In \emph{Proceedings of the 58th Annual Meeting of the Association for Computational Linguistics}, pages 4623--4637, Online. Association for Computational Linguistics.

\bibitem[{Artetxe and Schwenk(2019)}]{artetxe-schwenk-2019-massively}
Mikel Artetxe and Holger Schwenk. 2019.
\newblock \href {https://doi.org/10.1162/tacl_a_00288} {Massively multilingual sentence embeddings for zero-shot cross-lingual transfer and beyond}.
\newblock \emph{Transactions of the Association for Computational Linguistics}, 7:597--610.

\bibitem[{Belinkov(2022)}]{belinkov2022probing}
Yonatan Belinkov. 2022.
\newblock \href {https://doi.org/10.1162/coli_a_00422} {Probing classifiers: Promises, shortcomings, and advances}.
\newblock \emph{Computational Linguistics}, 48(1):207--219.

\bibitem[{Benjamini and Hochberg(1995)}]{benjamini1995controlling}
Yoav Benjamini and Yosef Hochberg. 1995.
\newblock Controlling the false discovery rate: a practical and powerful approach to multiple testing.
\newblock \emph{Journal of the Royal statistical society: series B (Methodological)}, 57(1):289--300.

\bibitem[{Choenni and Shutova(2020)}]{choenni2020what}
Rochelle Choenni and Ekaterina Shutova. 2020.
\newblock \href {http://arxiv.org/abs/2009.12862} {What does it mean to be language-agnostic? probing multilingual sentence encoders for typological properties}.
\newblock \emph{CoRR}, abs/2009.12862.

\bibitem[{Clark et~al.(2020)Clark, Choi, Collins, Garrette, Kwiatkowski, Nikolaev, and Palomaki}]{clark-etal-2020-tydi}
Jonathan~H. Clark, Eunsol Choi, Michael Collins, Dan Garrette, Tom Kwiatkowski, Vitaly Nikolaev, and Jennimaria Palomaki. 2020.
\newblock \href {https://doi.org/10.1162/tacl_a_00317} {{T}y{D}i {QA}: A benchmark for information-seeking question answering in typologically diverse languages}.
\newblock \emph{Transactions of the Association for Computational Linguistics}, 8:454--470.

\bibitem[{Conneau et~al.(2020)Conneau, Khandelwal, Goyal, Chaudhary, Wenzek, Guzm{\'a}n, Grave, Ott, Zettlemoyer, and Stoyanov}]{conneau-etal-2020-unsupervised}
Alexis Conneau, Kartikay Khandelwal, Naman Goyal, Vishrav Chaudhary, Guillaume Wenzek, Francisco Guzm{\'a}n, Edouard Grave, Myle Ott, Luke Zettlemoyer, and Veselin Stoyanov. 2020.
\newblock \href {https://doi.org/10.18653/v1/2020.acl-main.747} {Unsupervised cross-lingual representation learning at scale}.
\newblock In \emph{Proceedings of the 58th Annual Meeting of the Association for Computational Linguistics}, pages 8440--8451, Online. Association for Computational Linguistics.

\bibitem[{Conneau and Lample(2019{\natexlab{a}})}]{lample-2019-cross}
Alexis Conneau and Guillaume Lample. 2019{\natexlab{a}}.
\newblock \href {https://proceedings.neurips.cc/paper/2019/hash/c04c19c2c2474dbf5f7ac4372c5b9af1-Abstract.html} {Cross-lingual language model pretraining}.
\newblock In \emph{Advances in Neural Information Processing Systems 32: Annual Conference on Neural Information Processing Systems 2019, NeurIPS 2019, December 8-14, 2019, Vancouver, BC, Canada}, pages 7057--7067.

\bibitem[{Conneau and Lample(2019{\natexlab{b}})}]{NEURIPS2019_xlm}
Alexis Conneau and Guillaume Lample. 2019{\natexlab{b}}.
\newblock \href {https://proceedings.neurips.cc/paper/2019/hash/c04c19c2c2474dbf5f7ac4372c5b9af1-Abstract.html} {Cross-lingual language model pretraining}.
\newblock In \emph{Advances in Neural Information Processing Systems 32: Annual Conference on Neural Information Processing Systems 2019, NeurIPS 2019, December 8-14, 2019, Vancouver, BC, Canada}, pages 7057--7067.

\bibitem[{Conneau et~al.(2018)Conneau, Rinott, Lample, Williams, Bowman, Schwenk, and Stoyanov}]{conneau-etal-2018-xnli}
Alexis Conneau, Ruty Rinott, Guillaume Lample, Adina Williams, Samuel Bowman, Holger Schwenk, and Veselin Stoyanov. 2018.
\newblock \href {https://doi.org/10.18653/v1/D18-1269} {{XNLI}: Evaluating cross-lingual sentence representations}.
\newblock In \emph{Proceedings of the 2018 Conference on Empirical Methods in Natural Language Processing}, pages 2475--2485, Brussels, Belgium. Association for Computational Linguistics.

\bibitem[{Devlin(2019)}]{Devlin_2019}
Jacob Devlin. 2019.
\newblock \href {https://github.com/google-research/bert/blob/master/multilingual.md} {Multilingual {BERT} {README}}.

\bibitem[{Devlin et~al.(2019)Devlin, Chang, Lee, and Toutanova}]{devlin-etal-2019-bert}
Jacob Devlin, Ming-Wei Chang, Kenton Lee, and Kristina Toutanova. 2019.
\newblock \href {https://doi.org/10.18653/v1/N19-1423} {{BERT}: Pre-training of deep bidirectional transformers for language understanding}.
\newblock In \emph{Proceedings of the 2019 Conference of the North {A}merican Chapter of the Association for Computational Linguistics: Human Language Technologies, Volume 1 (Long and Short Papers)}, pages 4171--4186, Minneapolis, Minnesota. Association for Computational Linguistics.

\bibitem[{Dubossarsky et~al.(2020)Dubossarsky, Vuli{\'c}, Reichart, and Korhonen}]{dubossarsky-etal-2020-secret}
Haim Dubossarsky, Ivan Vuli{\'c}, Roi Reichart, and Anna Korhonen. 2020.
\newblock \href {https://doi.org/10.18653/v1/2020.emnlp-main.186} {The secret is in the spectra: Predicting cross-lingual task performance with spectral similarity measures}.
\newblock In \emph{Proceedings of the 2020 Conference on Empirical Methods in Natural Language Processing (EMNLP)}, pages 2377--2390, Online. Association for Computational Linguistics.

\bibitem[{Foroutan et~al.(2022)Foroutan, Banaei, Lebret, Bosselut, and Aberer}]{foroutan-etal-2022-discovering}
Negar Foroutan, Mohammadreza Banaei, R{\'e}mi Lebret, Antoine Bosselut, and Karl Aberer. 2022.
\newblock \href {https://aclanthology.org/2022.emnlp-main.513} {Discovering language-neutral sub-networks in multilingual language models}.
\newblock In \emph{Proceedings of the 2022 Conference on Empirical Methods in Natural Language Processing}, pages 7560--7575, Abu Dhabi, United Arab Emirates. Association for Computational Linguistics.

\bibitem[{Fujinuma et~al.(2022)Fujinuma, Boyd-Graber, and Kann}]{fujinuma-etal-2022-match}
Yoshinari Fujinuma, Jordan Boyd-Graber, and Katharina Kann. 2022.
\newblock \href {https://doi.org/10.18653/v1/2022.acl-long.106} {Match the script, adapt if multilingual: Analyzing the effect of multilingual pretraining on cross-lingual transferability}.
\newblock In \emph{Proceedings of the 60th Annual Meeting of the Association for Computational Linguistics (Volume 1: Long Papers)}, pages 1500--1512, Dublin, Ireland. Association for Computational Linguistics.

\bibitem[{Gerz et~al.(2018)Gerz, Vuli{\'c}, Ponti, Reichart, and Korhonen}]{gerz-etal-2018-relation}
Daniela Gerz, Ivan Vuli{\'c}, Edoardo~Maria Ponti, Roi Reichart, and Anna Korhonen. 2018.
\newblock \href {https://doi.org/10.18653/v1/D18-1029} {On the relation between linguistic typology and (limitations of) multilingual language modeling}.
\newblock In \emph{Proceedings of the 2018 Conference on Empirical Methods in Natural Language Processing}, pages 316--327, Brussels, Belgium. Association for Computational Linguistics.

\bibitem[{Ginter et~al.(2017)Ginter, Haji{\v c}, Luotolahti, Straka, and Zeman}]{ginter-etal-2017-conll}
Filip Ginter, Jan Haji{\v c}, Juhani Luotolahti, Milan Straka, and Daniel Zeman. 2017.
\newblock \href {http://hdl.handle.net/11234/1-1989} {{CoNLL} 2017 shared task - automatically annotated raw texts and word embeddings}.
\newblock {LINDAT}/{CLARIAH}-{CZ} digital library at the Institute of Formal and Applied Linguistics ({{\'U}FAL}), Faculty of Mathematics and Physics, Charles University.

\bibitem[{Harshman(1972b)}]{harshman-1972b-parafac2}
R.~A. Harshman. 1972b.
\newblock {P}{A}{R}{A}{F}{A}{C}2: {M}athematical and technical notes.
\newblock \emph{UCLA Working Papers in Phonetics}, 22:30--44.

\bibitem[{Hewitt and Manning(2019)}]{hewitt-manning-2019-structural}
John Hewitt and Christopher~D. Manning. 2019.
\newblock \href {https://doi.org/10.18653/v1/N19-1419} {{A} structural probe for finding syntax in word representations}.
\newblock In \emph{Proceedings of the 2019 Conference of the North {A}merican Chapter of the Association for Computational Linguistics: Human Language Technologies, Volume 1 (Long and Short Papers)}, pages 4129--4138, Minneapolis, Minnesota. Association for Computational Linguistics.

\bibitem[{Hu et~al.(2020)Hu, Ruder, Siddhant, Neubig, Firat, and Johnson}]{hu2020xtreme}
Junjie Hu, Sebastian Ruder, Aditya Siddhant, Graham Neubig, Orhan Firat, and Melvin Johnson. 2020.
\newblock \href {https://arxiv.org/abs/2003.11080} {Xtreme: A massively multilingual multi-task benchmark for evaluating cross-lingual generalization}.
\newblock \emph{ArXiv preprint}, abs/2003.11080.

\bibitem[{Jawahar et~al.(2019)Jawahar, Sagot, and Seddah}]{jawahar-etal-2019-bert}
Ganesh Jawahar, Beno{\^\i}t Sagot, and Djam{\'e} Seddah. 2019.
\newblock \href {https://doi.org/10.18653/v1/P19-1356} {What does {BERT} learn about the structure of language?}
\newblock In \emph{Proceedings of the 57th Annual Meeting of the Association for Computational Linguistics}, pages 3651--3657, Florence, Italy. Association for Computational Linguistics.

\bibitem[{K et~al.(2020)K, Wang, Mayhew, and Roth}]{karthikeyan-etal-2020-ability}
Karthikeyan K, Zihan Wang, Stephen Mayhew, and Dan Roth. 2020.
\newblock \href {https://openreview.net/forum?id=HJeT3yrtDr} {Cross-lingual ability of multilingual {BERT:} an empirical study}.
\newblock In \emph{8th International Conference on Learning Representations, {ICLR} 2020, Addis Ababa, Ethiopia, April 26-30, 2020}. OpenReview.net.

\bibitem[{Kendall(1948)}]{kendall1948rank}
Maurice~George Kendall. 1948.
\newblock \emph{Rank correlation methods.}
\newblock Griffin.

\bibitem[{Kirov et~al.(2018)Kirov, Cotterell, Sylak-Glassman, Walther, Vylomova, Xia, Faruqui, Mielke, McCarthy, K{\"u}bler, Yarowsky, Eisner, and Hulden}]{kirov-etal-2018-unimorph}
Christo Kirov, Ryan Cotterell, John Sylak-Glassman, G{\'e}raldine Walther, Ekaterina Vylomova, Patrick Xia, Manaal Faruqui, Sabrina~J. Mielke, Arya McCarthy, Sandra K{\"u}bler, David Yarowsky, Jason Eisner, and Mans Hulden. 2018.
\newblock \href {https://aclanthology.org/L18-1293} {{U}ni{M}orph 2.0: {U}niversal {M}orphology}.
\newblock In \emph{Proceedings of the Eleventh International Conference on Language Resources and Evaluation ({LREC} 2018)}, Miyazaki, Japan. European Language Resources Association (ELRA).

\bibitem[{Kleindessner et~al.(2023)Kleindessner, Donini, Russell, and Zafar}]{pmlr-v206-kleindessner23a}
Matth\"aus Kleindessner, Michele Donini, Chris Russell, and Muhammad~Bilal Zafar. 2023.
\newblock \href {https://proceedings.mlr.press/v206/kleindessner23a.html} {Efficient fair pca for fair representation learning}.
\newblock In \emph{Proceedings of The 26th International Conference on Artificial Intelligence and Statistics}, volume 206 of \emph{Proceedings of Machine Learning Research}, pages 5250--5270. PMLR.

\bibitem[{Kornblith et~al.(2019)Kornblith, Norouzi, Lee, and Hinton}]{kornblith-etal-2019-similarity}
Simon Kornblith, Mohammad Norouzi, Honglak Lee, and Geoffrey~E. Hinton. 2019.
\newblock \href {http://proceedings.mlr.press/v97/kornblith19a.html} {Similarity of neural network representations revisited}.
\newblock In \emph{Proceedings of the 36th International Conference on Machine Learning, {ICML} 2019, 9-15 June 2019, Long Beach, California, {USA}}, volume~97 of \emph{Proceedings of Machine Learning Research}, pages 3519--3529. {PMLR}.

\bibitem[{Kossaifi et~al.(2019)Kossaifi, Panagakis, Anandkumar, and Pantic}]{tensorly}
Jean Kossaifi, Yannis Panagakis, Anima Anandkumar, and Maja Pantic. 2019.
\newblock \href {http://jmlr.org/papers/v20/18-277.html} {Tensorly: Tensor learning in python}.
\newblock \emph{Journal of Machine Learning Research}, 20(26):1--6.

\bibitem[{Kudo and Richardson(2018)}]{kudo-richardson-2018-sentencepiece}
Taku Kudo and John Richardson. 2018.
\newblock \href {https://doi.org/10.18653/v1/D18-2012} {{S}entence{P}iece: A simple and language independent subword tokenizer and detokenizer for neural text processing}.
\newblock In \emph{Proceedings of the 2018 Conference on Empirical Methods in Natural Language Processing: System Demonstrations}, pages 66--71, Brussels, Belgium. Association for Computational Linguistics.

\bibitem[{Kudugunta et~al.(2019)Kudugunta, Bapna, Caswell, and Firat}]{kudugunta-etal-2019-investigating}
Sneha Kudugunta, Ankur Bapna, Isaac Caswell, and Orhan Firat. 2019.
\newblock \href {https://doi.org/10.18653/v1/D19-1167} {Investigating multilingual {NMT} representations at scale}.
\newblock In \emph{Proceedings of the 2019 Conference on Empirical Methods in Natural Language Processing and the 9th International Joint Conference on Natural Language Processing (EMNLP-IJCNLP)}, pages 1565--1575, Hong Kong, China. Association for Computational Linguistics.

\bibitem[{Lewis et~al.(2020)Lewis, Oguz, Rinott, Riedel, and Schwenk}]{lewis-etal-2020-mlqa}
Patrick Lewis, Barlas Oguz, Ruty Rinott, Sebastian Riedel, and Holger Schwenk. 2020.
\newblock \href {https://doi.org/10.18653/v1/2020.acl-main.653} {{MLQA}: Evaluating cross-lingual extractive question answering}.
\newblock In \emph{Proceedings of the 58th Annual Meeting of the Association for Computational Linguistics}, pages 7315--7330, Online. Association for Computational Linguistics.

\bibitem[{Liang et~al.(2021)Liang, Dufter, and Sch{\"{u}}tze}]{liang2021locating}
Sheng Liang, Philipp Dufter, and Hinrich Sch{\"{u}}tze. 2021.
\newblock \href {http://arxiv.org/abs/2109.08040} {Locating language-specific information in contextualized embeddings}.
\newblock \emph{CoRR}, abs/2109.08040.

\bibitem[{Libovick{\'{y}} et~al.(2019)Libovick{\'{y}}, Rosa, and Fraser}]{Libovick-2019-HowLI}
Jindrich Libovick{\'{y}}, Rudolf Rosa, and Alexander Fraser. 2019.
\newblock \href {https://arxiv.org/abs/1911.03310} {How language-neutral is multilingual bert?}
\newblock \emph{ArXiv preprint}, abs/1911.03310.

\bibitem[{Liu et~al.(2020)Liu, Zhang, Zhang, Singh, Saraf, and Zweig}]{liu-etal-2020-multilingual}
Chunxi Liu, Qiaochu Zhang, Xiaohui Zhang, Kritika Singh, Yatharth Saraf, and Geoffrey Zweig. 2020.
\newblock \href {https://aclanthology.org/2020.sltu-1.7} {Multilingual graphemic hybrid {ASR} with massive data augmentation}.
\newblock In \emph{Proceedings of the 1st Joint Workshop on Spoken Language Technologies for Under-resourced languages (SLTU) and Collaboration and Computing for Under-Resourced Languages (CCURL)}, pages 46--52, Marseille, France. European Language Resources association.

\bibitem[{Liu et~al.(2019)Liu, Ott, Goyal, Du, Joshi, Chen, Levy, Lewis, Zettlemoyer, and Stoyanov}]{Liu2019RoBERTaAR}
Yinhan Liu, Myle Ott, Naman Goyal, Jingfei Du, Mandar Joshi, Danqi Chen, Omer Levy, Mike Lewis, Luke Zettlemoyer, and Veselin Stoyanov. 2019.
\newblock \href {https://arxiv.org/abs/1907.11692} {Roberta: A robustly optimized bert pretraining approach}.
\newblock \emph{ArXiv preprint}, abs/1907.11692.

\bibitem[{Loshchilov and Hutter(2019)}]{loshchilov-hutter-2019-decoupled}
Ilya Loshchilov and Frank Hutter. 2019.
\newblock \href {https://openreview.net/forum?id=Bkg6RiCqY7} {Decoupled weight decay regularization}.
\newblock In \emph{7th International Conference on Learning Representations, {ICLR} 2019, New Orleans, LA, USA, May 6-9, 2019}. OpenReview.net.

\bibitem[{Mann(1945)}]{mann1945nonparametric}
Henry~B Mann. 1945.
\newblock Nonparametric tests against trend.
\newblock \emph{Econometrica: Journal of the econometric society}, pages 245--259.

\bibitem[{McCarthy et~al.(2018)McCarthy, Silfverberg, Cotterell, Hulden, and Yarowsky}]{mccarthy-etal-2018-marrying}
Arya~D. McCarthy, Miikka Silfverberg, Ryan Cotterell, Mans Hulden, and David Yarowsky. 2018.
\newblock \href {https://doi.org/10.18653/v1/W18-6011} {Marrying {U}niversal {D}ependencies and {U}niversal {M}orphology}.
\newblock In \emph{Proceedings of the Second Workshop on Universal Dependencies ({UDW} 2018)}, pages 91--101, Brussels, Belgium. Association for Computational Linguistics.

\bibitem[{Merchant et~al.(2020)Merchant, Rahimtoroghi, Pavlick, and Tenney}]{merchant-etal-2020-happens}
Amil Merchant, Elahe Rahimtoroghi, Ellie Pavlick, and Ian Tenney. 2020.
\newblock \href {https://doi.org/10.18653/v1/2020.blackboxnlp-1.4} {What happens to {BERT} embeddings during fine-tuning?}
\newblock In \emph{Proceedings of the Third BlackboxNLP Workshop on Analyzing and Interpreting Neural Networks for NLP}, pages 33--44, Online. Association for Computational Linguistics.

\bibitem[{Morcos et~al.(2018)Morcos, Raghu, and Bengio}]{morcos-etal-2018-insights}
Ari~S. Morcos, Maithra Raghu, and Samy Bengio. 2018.
\newblock \href {https://proceedings.neurips.cc/paper/2018/hash/a7a3d70c6d17a73140918996d03c014f-Abstract.html} {Insights on representational similarity in neural networks with canonical correlation}.
\newblock In \emph{Advances in Neural Information Processing Systems 31: Annual Conference on Neural Information Processing Systems 2018, NeurIPS 2018, December 3-8, 2018, Montr{\'{e}}al, Canada}, pages 5732--5741.

\bibitem[{Muller et~al.(2021)Muller, Elazar, Sagot, and Seddah}]{muller-etal-2021-first}
Benjamin Muller, Yanai Elazar, Beno{\^\i}t Sagot, and Djam{\'e} Seddah. 2021.
\newblock \href {https://doi.org/10.18653/v1/2021.eacl-main.189} {First align, then predict: Understanding the cross-lingual ability of multilingual {BERT}}.
\newblock In \emph{Proceedings of the 16th Conference of the European Chapter of the Association for Computational Linguistics: Main Volume}, pages 2214--2231, Online. Association for Computational Linguistics.

\bibitem[{Nivre et~al.(2017{\natexlab{a}})Nivre, Zeman, Ginter, and Tyers}]{nivre-2017-ud}
Joakim Nivre, Daniel Zeman, Filip Ginter, and Francis Tyers. 2017{\natexlab{a}}.
\newblock \href {https://aclanthology.org/E17-5001} {{U}niversal {D}ependencies}.
\newblock In \emph{Proceedings of the 15th Conference of the {E}uropean Chapter of the Association for Computational Linguistics: Tutorial Abstracts}, Valencia, Spain. Association for Computational Linguistics.

\bibitem[{Nivre et~al.(2017{\natexlab{b}})Nivre, Zeman, Ginter, and Tyers}]{nivre-etal-udv2.2-2018}
Joakim Nivre, Daniel Zeman, Filip Ginter, and Francis Tyers. 2017{\natexlab{b}}.
\newblock \href {https://aclanthology.org/E17-5001} {{U}niversal {D}ependencies}.
\newblock In \emph{Proceedings of the 15th Conference of the {E}uropean Chapter of the Association for Computational Linguistics: Tutorial Abstracts}, Valencia, Spain. Association for Computational Linguistics.

\bibitem[{Olfat and Aswani(2019)}]{olfat2019convex}
Matt Olfat and Anil Aswani. 2019.
\newblock Convex formulations for fair principal component analysis.
\newblock In \emph{Proceedings of the AAAI Conference on Artificial Intelligence}, volume~33, pages 663--670.

\bibitem[{Pan et~al.(2017)Pan, Zhang, May, Nothman, Knight, and Ji}]{pan-etal-2017-cross}
Xiaoman Pan, Boliang Zhang, Jonathan May, Joel Nothman, Kevin Knight, and Heng Ji. 2017.
\newblock \href {https://doi.org/10.18653/v1/P17-1178} {Cross-lingual name tagging and linking for 282 languages}.
\newblock In \emph{Proceedings of the 55th Annual Meeting of the Association for Computational Linguistics (Volume 1: Long Papers)}, pages 1946--1958, Vancouver, Canada. Association for Computational Linguistics.

\bibitem[{Paszke et~al.(2019)Paszke, Gross, Massa, Lerer, Bradbury, Chanan, Killeen, Lin, Gimelshein, Antiga, Desmaison, K{\"{o}}pf, Yang, DeVito, Raison, Tejani, Chilamkurthy, Steiner, Fang, Bai, and Chintala}]{NEURIPS2019_9015}
Adam Paszke, Sam Gross, Francisco Massa, Adam Lerer, James Bradbury, Gregory Chanan, Trevor Killeen, Zeming Lin, Natalia Gimelshein, Luca Antiga, Alban Desmaison, Andreas K{\"{o}}pf, Edward Yang, Zachary DeVito, Martin Raison, Alykhan Tejani, Sasank Chilamkurthy, Benoit Steiner, Lu~Fang, Junjie Bai, and Soumith Chintala. 2019.
\newblock \href {https://proceedings.neurips.cc/paper/2019/hash/bdbca288fee7f92f2bfa9f7012727740-Abstract.html} {Pytorch: An imperative style, high-performance deep learning library}.
\newblock In \emph{Advances in Neural Information Processing Systems 32: Annual Conference on Neural Information Processing Systems 2019, NeurIPS 2019, December 8-14, 2019, Vancouver, BC, Canada}, pages 8024--8035.

\bibitem[{Pimentel et~al.(2020)Pimentel, Valvoda, Maudslay, Zmigrod, Williams, and Cotterell}]{pimentel-etal-2020-information}
Tiago Pimentel, Josef Valvoda, Rowan~Hall Maudslay, Ran Zmigrod, Adina Williams, and Ryan Cotterell. 2020.
\newblock \href {https://doi.org/10.18653/v1/2020.acl-main.420} {Information-theoretic probing for linguistic structure}.
\newblock In \emph{Proceedings of the 58th Annual Meeting of the Association for Computational Linguistics}, pages 4609--4622, Online. Association for Computational Linguistics.

\bibitem[{Qiu et~al.(2023)Qiu, Ziser, Korhonen, Ponti, and Cohen}]{qiu2023detecting}
Yifu Qiu, Yftah Ziser, Anna Korhonen, Edoardo~M. Ponti, and Shay~B. Cohen. 2023.
\newblock \href {http://arxiv.org/abs/2305.13632} {Detecting and mitigating hallucinations in multilingual summarisation}.

\bibitem[{Raghu et~al.(2017)Raghu, Gilmer, Yosinski, and Sohl{-}Dickstein}]{raghu-etal-2017-svcca}
Maithra Raghu, Justin Gilmer, Jason Yosinski, and Jascha Sohl{-}Dickstein. 2017.
\newblock \href {https://proceedings.neurips.cc/paper/2017/hash/dc6a7e655d7e5840e66733e9ee67cc69-Abstract.html} {{SVCCA:} singular vector canonical correlation analysis for deep learning dynamics and interpretability}.
\newblock In \emph{Advances in Neural Information Processing Systems 30: Annual Conference on Neural Information Processing Systems 2017, December 4-9, 2017, Long Beach, CA, {USA}}, pages 6076--6085.

\bibitem[{Rogers et~al.(2020)Rogers, Kovaleva, and Rumshisky}]{rogers-etal-2020-primer}
Anna Rogers, Olga Kovaleva, and Anna Rumshisky. 2020.
\newblock \href {https://doi.org/10.1162/tacl_a_00349} {A primer in {BERT}ology: What we know about how {BERT} works}.
\newblock \emph{Transactions of the Association for Computational Linguistics}, 8:842--866.

\bibitem[{Shao et~al.(2023{\natexlab{a}})Shao, Ziser, and Cohen}]{10.1162/tacl_a_00558}
Shun Shao, Yftah Ziser, and Shay~B. Cohen. 2023{\natexlab{a}}.
\newblock \href {https://doi.org/10.1162/tacl_a_00558} {{Erasure of Unaligned Attributes from Neural Representations}}.
\newblock \emph{Transactions of the Association for Computational Linguistics}, 11:488--510.

\bibitem[{Shao et~al.(2023{\natexlab{b}})Shao, Ziser, and Cohen}]{shao2022gold}
Shun Shao, Yftah Ziser, and Shay~B. Cohen. 2023{\natexlab{b}}.
\newblock \href {https://aclanthology.org/2023.eacl-main.118} {Gold doesn{'}t always glitter: Spectral removal of linear and nonlinear guarded attribute information}.
\newblock In \emph{Proceedings of the 17th Conference of the European Chapter of the Association for Computational Linguistics}, pages 1611--1622, Dubrovnik, Croatia. Association for Computational Linguistics.

\bibitem[{Singh et~al.(2019)Singh, McCann, Socher, and Xiong}]{singh-etal-2019-bert}
Jasdeep Singh, Bryan McCann, Richard Socher, and Caiming Xiong. 2019.
\newblock \href {https://doi.org/10.18653/v1/D19-6106} {{BERT} is not an interlingua and the bias of tokenization}.
\newblock In \emph{Proceedings of the 2nd Workshop on Deep Learning Approaches for Low-Resource NLP (DeepLo 2019)}, pages 47--55, Hong Kong, China. Association for Computational Linguistics.

\bibitem[{Sokal and Michener(1958)}]{Sokal-1958-ASM}
Robert~R. Sokal and Charles~Duncan Michener. 1958.
\newblock A statistical method for evaluating systematic relationships.
\newblock \emph{University of Kansas science bulletin}, 38:1409--1438.

\bibitem[{Stanczak et~al.(2022)Stanczak, Ponti, Torroba~Hennigen, Cotterell, and Augenstein}]{stanczak-etal-2022-neurons}
Karolina Stanczak, Edoardo Ponti, Lucas Torroba~Hennigen, Ryan Cotterell, and Isabelle Augenstein. 2022.
\newblock \href {https://doi.org/10.18653/v1/2022.naacl-main.114} {Same neurons, different languages: Probing morphosyntax in multilingual pre-trained models}.
\newblock In \emph{Proceedings of the 2022 Conference of the North American Chapter of the Association for Computational Linguistics: Human Language Technologies}, pages 1589--1598, Seattle, United States. Association for Computational Linguistics.

\bibitem[{Tenney et~al.(2019)Tenney, Das, and Pavlick}]{tenney-etal-2019-bert}
Ian Tenney, Dipanjan Das, and Ellie Pavlick. 2019.
\newblock \href {https://doi.org/10.18653/v1/P19-1452} {{BERT} rediscovers the classical {NLP} pipeline}.
\newblock In \emph{Proceedings of the 57th Annual Meeting of the Association for Computational Linguistics}, pages 4593--4601, Florence, Italy. Association for Computational Linguistics.

\bibitem[{Torroba~Hennigen et~al.(2020)Torroba~Hennigen, Williams, and Cotterell}]{torroba-hennigen-etal-2020-intrinsic}
Lucas Torroba~Hennigen, Adina Williams, and Ryan Cotterell. 2020.
\newblock \href {https://doi.org/10.18653/v1/2020.emnlp-main.15} {Intrinsic probing through dimension selection}.
\newblock In \emph{Proceedings of the 2020 Conference on Empirical Methods in Natural Language Processing (EMNLP)}, pages 197--216, Online. Association for Computational Linguistics.

\bibitem[{Wolf et~al.(2020)Wolf, Debut, Sanh, Chaumond, Delangue, Moi, Cistac, Rault, Louf, Funtowicz, Davison, Shleifer, von Platen, Ma, Jernite, Plu, Xu, Le~Scao, Gugger, Drame, Lhoest, and Rush}]{wolf2019huggingface}
Thomas Wolf, Lysandre Debut, Victor Sanh, Julien Chaumond, Clement Delangue, Anthony Moi, Pierric Cistac, Tim Rault, Remi Louf, Morgan Funtowicz, Joe Davison, Sam Shleifer, Patrick von Platen, Clara Ma, Yacine Jernite, Julien Plu, Canwen Xu, Teven Le~Scao, Sylvain Gugger, Mariama Drame, Quentin Lhoest, and Alexander Rush. 2020.
\newblock \href {https://doi.org/10.18653/v1/2020.emnlp-demos.6} {Transformers: State-of-the-art natural language processing}.
\newblock In \emph{Proceedings of the 2020 Conference on Empirical Methods in Natural Language Processing: System Demonstrations}, pages 38--45, Online. Association for Computational Linguistics.

\bibitem[{Xue et~al.(2021)Xue, Constant, Roberts, Kale, Al-Rfou, Siddhant, Barua, and Raffel}]{xue-etal-2021-mt5}
Linting Xue, Noah Constant, Adam Roberts, Mihir Kale, Rami Al-Rfou, Aditya Siddhant, Aditya Barua, and Colin Raffel. 2021.
\newblock \href {https://doi.org/10.18653/v1/2021.naacl-main.41} {m{T}5: A massively multilingual pre-trained text-to-text transformer}.
\newblock In \emph{Proceedings of the 2021 Conference of the North American Chapter of the Association for Computational Linguistics: Human Language Technologies}, pages 483--498, Online. Association for Computational Linguistics.

\bibitem[{Yang et~al.(2019)Yang, Zhang, Tar, and Baldridge}]{yang-etal-2019-paws}
Yinfei Yang, Yuan Zhang, Chris Tar, and Jason Baldridge. 2019.
\newblock \href {https://doi.org/10.18653/v1/D19-1382} {{PAWS}-{X}: A cross-lingual adversarial dataset for paraphrase identification}.
\newblock In \emph{Proceedings of the 2019 Conference on Empirical Methods in Natural Language Processing and the 9th International Joint Conference on Natural Language Processing (EMNLP-IJCNLP)}, pages 3687--3692, Hong Kong, China. Association for Computational Linguistics.

\bibitem[{Zhao et~al.(2022)Zhao, Ziser, and Cohen}]{zhao-etal-2022-understanding}
Zheng Zhao, Yftah Ziser, and Shay Cohen. 2022.
\newblock \href {https://aclanthology.org/2022.blackboxnlp-1.16} {Understanding domain learning in language models through subpopulation analysis}.
\newblock In \emph{Proceedings of the Fifth BlackboxNLP Workshop on Analyzing and Interpreting Neural Networks for NLP}, pages 192--209, Abu Dhabi, United Arab Emirates (Hybrid). Association for Computational Linguistics.

\bibitem[{Ziser and Reichart(2018)}]{ziser-reichart-2018-deep}
Yftah Ziser and Roi Reichart. 2018.
\newblock \href {https://doi.org/10.18653/v1/D18-1022} {Deep pivot-based modeling for cross-language cross-domain transfer with minimal guidance}.
\newblock In \emph{Proceedings of the 2018 Conference on Empirical Methods in Natural Language Processing}, pages 238--249, Brussels, Belgium. Association for Computational Linguistics.

\bibitem[{Zweigenbaum et~al.(2017)Zweigenbaum, Sharoff, and Rapp}]{zweigenbaum-etal-2017-overview}
Pierre Zweigenbaum, Serge Sharoff, and Reinhard Rapp. 2017.
\newblock \href {https://doi.org/10.18653/v1/W17-2512} {Overview of the second {BUCC} shared task: Spotting parallel sentences in comparable corpora}.
\newblock In \emph{Proceedings of the 10th Workshop on Building and Using Comparable Corpora}, pages 60--67, Vancouver, Canada. Association for Computational Linguistics.

\end{thebibliography}
\bibliographystyle{acl_natbib}

\clearpage
\newpage

\appendix

\section{Information on Attributes and Languages}
\label{app:appendix}

We first provide information about all languages we use in our experiment in Table~\ref{tab:languages_info}. The information includes ISO 639-1 codes for all languages, the language family and the genus they belong to. In Table~\ref{tab:languages_morpho}, we present all morphosyntactic attributes we experiment. For each attribute, we list all languages that have the attribute. We also provide a reverse list where we list by languages:
\begin{itemize}
   \item \textbf{Arabic (ar)}: Aspect, Case, Definiteness, Finiteness, Gender, Mood, Number, Part of Speech, Person, Polarity, Politeness, Voice 
   \item \textbf{Bulgarian (bg)}: Aspect, Case, Comparison, Definiteness, Gender, Mood, Number, Part of Speech, Person, Polarity, Tense, Valency, Voice
   \item \textbf{Catalan (ca)}: Aspect, Case, Definiteness, Finiteness, Gender, Mood, Number, Part of Speech, Person, Polarity, Possession, Tense
   \item \textbf{Chinese (zh)}: Aspect, Case, Number, Part of Speech, Person, Polarity, Valency, Voice
   \item \textbf{Croatian (hr)}: Animacy, Case, Comparison, Definiteness, Finiteness, Gender, Mood, Number, Part of Speech, Person, Polarity, Possession, Tense, Valency, Voice
   \item \textbf{Czech (cs)}: Animacy, Aspect, Case, Comparison, Finiteness, Gender, Mood, Number, Part of Speech, Person, Polarity, Possession, Tense, Valency, Voice
   \item \textbf{Danish (da)}: Case, Comparison, Definiteness, Finiteness, Gender, Mood, Number, Part of Speech, Person, Possession, Tense, Valency, Voice  
   \item \textbf{Dutch (nl)}: Case, Comparison, Definiteness, Finiteness, Gender, Number, Part of Speech, Person, Tense, Valency
   \item \textbf{English (en)}: Case, Comparison, Definiteness, Finiteness, Gender, Mood, Number, Part of Speech, Person, Tense, Valency
   \item \textbf{Estonian (et)}: Aspect, Case, Comparison, Finiteness, Mood, Number, Part of Speech, Person, Polarity, Tense, Valency, Voice 
   \item \textbf{Finnish (fi)}: Case, Comparison, Finiteness, Mood, Number, Part of Speech, Person, Polarity, Possession, Tense, Valency, Voice
   \item \textbf{French (fr)}: Aspect, Definiteness, Finiteness, Gender, Mood, Number, Part of Speech, Person, Polarity, Tense, Valency
   \item \textbf{Galician (gl)}: Part of Speech, Polarity  
   \item \textbf{German (de)}: Case, Comparison, Definiteness, Finiteness, Mood, Number, Part of Speech, Person, Polarity, Politeness, Possession, Tense, Valency
   \item \textbf{Greek (el)}: Aspect, Case, Comparison, Definiteness, Finiteness, Gender, Mood, Number, Part of Speech, Person, Tense, Voice 
   \item \textbf{Hebrew (he)}: Case, Definiteness, Finiteness, Number, Part of Speech, Person, Polarity, Possession, Tense, Valency, Voice
   \item \textbf{Hindi (hi)}: Aspect, Case, Finiteness, Gender, Mood, Number, Part of Speech, Person, Polarity, Politeness, Tense, Voice
   \item \textbf{Hungarian (hu)}: Case, Comparison, Definiteness, Finiteness, Mood, Number, Part of Speech, Person, Possession, Tense, Valency    
   \item \textbf{Indonesian (id)}: Part of Speech, Polarity  
   \item \textbf{Italian (it)}: Aspect, Comparison, Definiteness, Finiteness, Gender, Mood, Number, Part of Speech, Person, Tense
   \item \textbf{Japanese (ja)}: Part of Speech
   \item \textbf{Korean (ko)}: Part of Speech
   \item \textbf{Polish (pl)}: Animacy, Aspect, Case, Comparison, Finiteness, Gender, Mood, Number, Part of Speech, Person, Polarity, Possession, Tense, Valency, Voice
   \item \textbf{Portuguese (pt)}: Aspect, Case, Definiteness, Finiteness, Gender, Mood, Number, Part of Speech, Person, Polarity, Tense
   \item \textbf{Romanian (ro)}: Aspect, Case, Definiteness, Finiteness, Gender, Mood, Number, Part of Speech, Person, Polarity, Possession, Tense, Valency
   \item \textbf{Russian (ru)}: Animacy, Aspect, Case, Comparison, Finiteness, Gender, Mood, Number, Part of Speech, Person, Polarity, Tense, Valency, Voice
   \item \textbf{Slovak  (sk)}: Animacy, Aspect, Case, Comparison, Finiteness, Gender, Mood, Number, Part of Speech, Person, Polarity, Possession, Tense, Valency, Voice
   \item \textbf{Slovenian (sl)}: Animacy, Aspect, Case, Comparison, Definiteness, Finiteness, Gender, Mood, Number, Part of Speech, Person, Polarity, Possession, Tense, Valency  
   \item \textbf{Spanish (es)}: Aspect, Case, Comparison, Definiteness, Finiteness, Gender, Mood, Number, Part of Speech, Person, Polarity, Tense, Valency
   \item \textbf{Swedish (sv)}: Case, Comparison, Definiteness, Finiteness, Gender, Mood, Number, Part of Speech, Polarity, Tense, Voice
   \item \textbf{Turkish (tr)}: Aspect, Case, Mood, Number, Part of Speech, Person, Polarity, Politeness, Possession, Tense, Valency, Voice
   \item \textbf{Ukrainian (uk)}: Animacy, Aspect, Case, Comparison, Finiteness, Gender, Mood, Number, Part of Speech, Person, Polarity, Tense, Valency, Voice    
   \item \textbf{Vietnamese (vi)}: Part of Speech, Polarity  
\end{itemize}
Notice that in this list and in our work, we omit a language that has less than 100 instances labeled for a particular morphosyntactic category. 

\begin{table*}[ht]
\centering
\begin{tabular}{lclr}
\Xhline{1pt}
\textbf{Language} & \textbf{ISO 639-1} & \textbf{Family} & \textbf{Genus} \\
\hline
   Arabic      &   ar      &   Afro-Asiatic     &   Semitic \\
   Bulgarian   &   bg      &   Indo-European     &   Slavic \\
   Catalan     &   ca      &   Indo-European     &  Romance \\
   Chinese     &   zh      &   Sino-Tibetan     &   Chinese \\ 
   Croatian    &   hr      &   Indo-European     &   Slavic \\
   Czech       &   cs      &   Indo-European     &   Slavic \\
   Danish      &   da      &   Indo-European     &   Germanic \\
   Dutch       &   nl      &   Indo-European     &   Germanic \\
   English     &   en      &   Indo-European   &    Germanic \\
   Estonian    &   et      &   Uralic      &   Finnic \\
   Finnish     &   fi      &   Uralic     &   Finnic \\
   French      &   fr      &   Indo-European   &   Romance \\
   Galician    &   gl      &   Indo-European     & Romance \\
   German      &   de      &   Indo-European   &   Germanic \\ 
   Greek       &   el      &   Indo-European   &   Greek \\
   Hebrew      &   he      &   Afro-Asiatic   &   Semitic   \\
   Hindi       &   hi      &   Indo-European      &   Indic \\ 
   Hungarian   &   hu      &   Uralic     &   Ugric  \\
   Indonesian  &   id      &   Austronesian    &   Malayo-Sumbawan \\
   Italian     &   it      &   Indo-European   &   Romance \\
   Japanese    &   ja      &   Japanese      &   Japanese \\
   Korean      &   ko      &   Korean     &   Korean \\
   Polish      &   pl      &   Indo-European     &   Slavic \\
   Portuguese  &   pt      &   Indo-European     &   Romance \\
   Romanian    &   ro      &   Indo-European     &  Romance \\
   Russian     &   ru      &   Indo-European   &   Slavic \\
   Slovak      &   sk      &   Indo-European     & Slavic \\
   Slovenian   &   sl      &   Indo-European     & Slavic \\
   Spanish     &   es      &   Indo-European   &   Romance \\
   Swedish     &   sv      &   Indo-European   &   Germanic \\
   Turkish     &   tr      &   Altaic     &   Turkic \\
   Ukrainian   &   uk      &   Indo-European     &   Slavic \\
   Vietnamese  &   vi      &   Austro-Asiatic      & Vietic \\

\Xhline{1pt}
\end{tabular}
\caption{All languages used in this work along with language family and genus they belong to. Family and genus information is obtained from WALS.}
\label{tab:languages_info}
\end{table*}

\begin{table*}[ht]
\small
\centering
\begin{tabular}{ll}
\Xhline{1pt}
\textbf{Property} & \textbf{Language}  \\
\hline
{Animacy} & bg, hr, cs, pl, ru, sk, sl, uk, \\

{Aspect} & ar, bg, ca, zh, hr, cs, et, fr, el, hi, hu, it, 
     pl, pt, ro, ru, sk, sl, es, tr, uk \\

{Case} & ar, bg, ca, zh, hr, cs, da, nl, en, et, 
    fi, fr, de, el, he, hi, hu,
    pl, pt, ro, ru, sk, sl, es, sv, tr, uk \\

{Comparison} & bg, hr, cs, da, nl, en, et, 
    fi, fr, de, el, hu, it, 
    pl, ro, ru, sk, sl, es, sv, uk \\

{Definiteness} & ar, bg, ca, hr, da, nl, en, 
    fr, de, el, he, hu, it, 
    pt, ro, sl, es, sv \\

{Finiteness} & ar, ca, hr, cs, da, nl, en, et, 
    fi, fr, de, el, he, hi, hu, it, 
    pl, pt, ro, ru, sk, sl, es, sv, uk \\

{Gender} & ar, bg, ca, hr, cs, da, nl, en,
    fr, el, hi, it, 
    pl, pt, ro, ru, sk, sl, es, sv, uk \\
    
{Mood} & ar, bg, ca, hr, cs, da, en, et, 
    fi, fr, de, el, he, hi, hu, it, 
    pl, pt, ro, ru, sk, sl, es, sv, tr, uk \\

{Number} & ar, bg, ca, zh, hr, cs, da, nl, en, et, 
    fi, fr, de, el, he, hi, hu, it, 
    pl, pt, ro, ru, sk, sl, es, sv, tr, uk \\
    
{Part of Speech} & all 33 languages \\

{Person} & ar, bg, ca, zh, hr, cs, da, nl, en, et, 
    fi, fr, de, el, he, hi, hu, it, 
    pl, pt, ro, ru, sk, sl, es, tr, uk \\

{Polarity} & ar, bg, ca, zh, hr, cs, et, 
    fi, fr, gl, de, he, hi, id, it, 
    pl, pt, ro, ru, sk, sl, es, sv, tr, uk, vi \\

{Politeness} & ca, da, de, hi, es, tr \\
    
{Possession} & ar, ca, hr, cs, da, fi, de, he, hu, pl, ro, sk, sl, tr \\

{Tense} & bg, ca, hr, cs, da, nl, en, et, 
    fi, fr, de, el, he, hi, hu, it, 
    pl, pt, ro, ru, sk, sl, es, sv, tr, uk \\

{Valency} & bg, zh, hr, cs, da, nl, en, et, 
    fi, fr, de, he, hu,
    pl, ro, ru, sk, sl, es, tr, uk \\

{Voice} & ar, bg, zh, hr, cs, da, et, 
    fi, el, he, hi, 
    pl, ru, sk, sv, tr, uk \\
\Xhline{1pt}
\end{tabular}
\caption{All language properties (morphosyntactic categories) we analyzed in this work along with languages that have the corresponding property.}
\label{tab:languages_morpho}
\end{table*}

\section{Additional Details for Experiments}
\subsection{Details for Data}
\label{app:exp_data}
We use CoNLL's 2017 Wikipedia dump \citep{ginter-etal-2017-conll} to train our models. Following \citet{fujinuma-etal-2022-match}, we downsample all Wikipedia datasets to an identical number of sequences in order to use the same amount of data for all language pre-training. The downsampled dataset is standardized to the Hindi corpus, which has the smallest size among all languages we examine. For each language's pre-training data, there are about 30M tokens (approximately 200MB). In total, we experiment with 33 languages. We provide the full list of languages used for our experiments in Appendix~\ref{app:appendix}. We also create a validation set with 1K sequences (about 512 tokens per sequence) to measure model loss (cross-entropy) during pre-training. For morphosyntactic features, we use treebanks from UD 2.1 \citep{nivre-2017-ud}. These treebanks contain sentences annotated with morphosyntactic information and are available for a wide range of languages. We obtain a contextual representation for every word in the treebanks by feeding them to our multilingual/monolingual models. We then use the UniMorph schema \citep{kirov-etal-2018-unimorph} to map each word with its parts of speech and morphosyntactic properties. We provide a list of morphosyntactic categories we use in Appendix~\ref{app:appendix}. We follow \citet{stanczak-etal-2022-neurons} and use the converter \citep{mccarthy-etal-2018-marrying} to switch morphosyntactic annotations from UD v2.1 to UniMorph schema.

\subsection{Details for Models}
\label{app:exp_model}
We use the XLM-R \citep{conneau-etal-2020-unsupervised} architecture for the multilingual $\mymat{E}$ model, and we use RoBERTa \citep{Liu2019RoBERTaAR} for the monolingual $\mymat{C}_{\ell}$ model. We use the base variant for both kinds of models, which consists of 12 layers, 768 hidden dimensions, 8 attention heads for RoBERTa, and 12 attention heads for XLM-R. We use XLM-R's vocabulary and the SentencePiece \citep{kudo-richardson-2018-sentencepiece} tokenizer for all our experiments provided by \citet{conneau-etal-2020-unsupervised} in order to support all languages we analyze and enable fair comparison for all configurations. We do not use the original RoBERTa vocabulary and tokenizer since they only support English. We pre-train all models for a maximum of 150K steps, and all models use the validation set cross-entropy loss to perform early stopping. We use the AdamW optimizer \citep{loshchilov-hutter-2019-decoupled} with a learning rate of $10^{-4}$. Our monolingual models were trained on four NVIDIA GeForce GTX 1080 Ti GPUs with a batch size of two per GPU, and our multilingual models were trained on four NVIDIA A100 GPUs with a batch size of 16 per GPU. Both models take about two days to train. We use PyTorch \citep{NEURIPS2019_9015}, the HuggingFace library \citep{wolf2019huggingface} and the TensorLy library \citep{tensorly} for all model implementation and PARAFAC2 computation.

\section{Additional Results for RQ1}
\label{app:rq1}

\begin{figure}[ht]
     \centering
    \includegraphics[width=\linewidth]{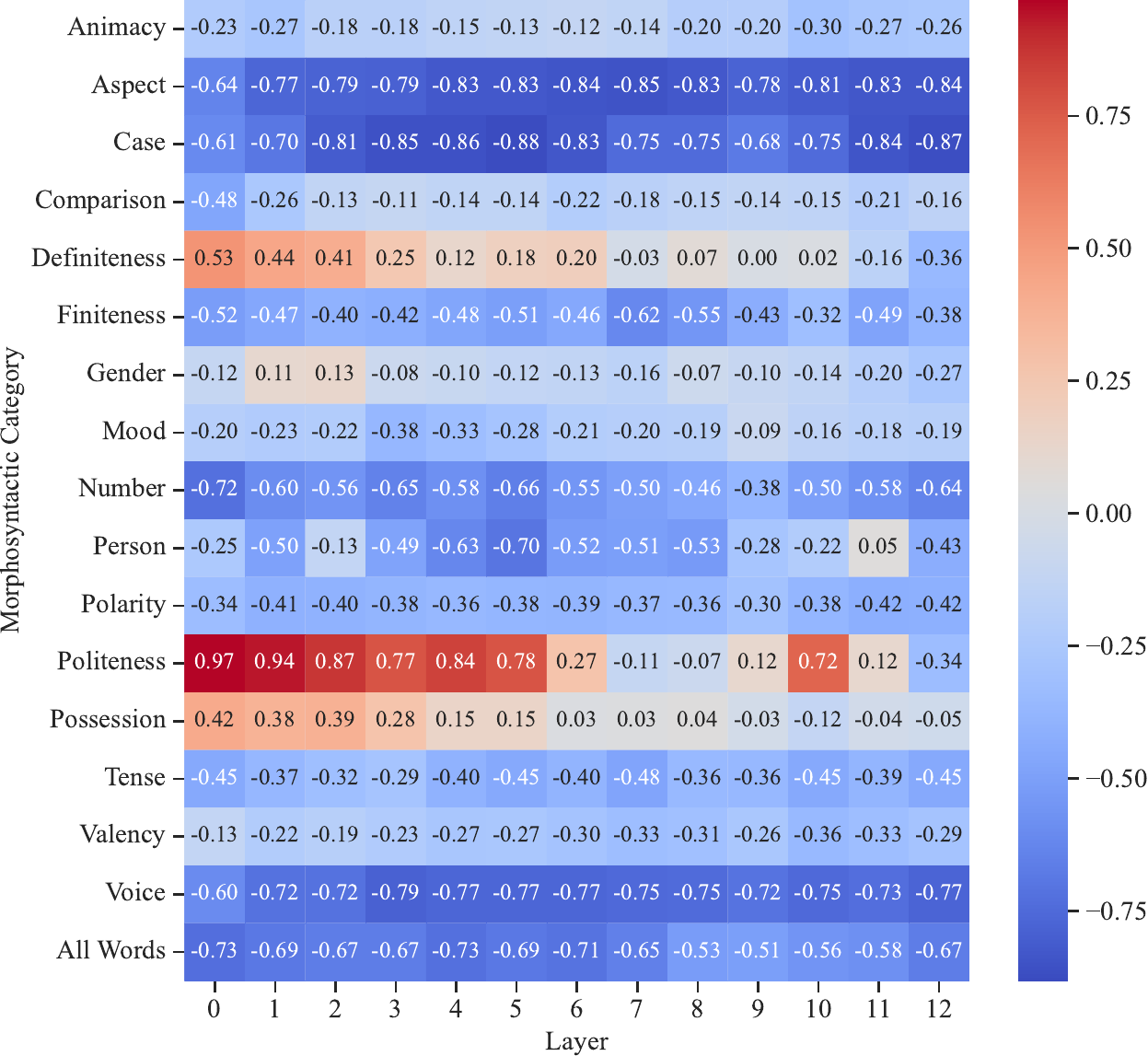}       
\caption{Pearson correlation results between the average $\mathrm{sig}(\ell)$ for all languages and their number of unique characters for each morphosyntactic category among all layers.}
\label{fig:correlation_uniq_chars_all_layers}
\end{figure}

\begin{figure}[ht]
     \centering
    \includegraphics[width=\linewidth]{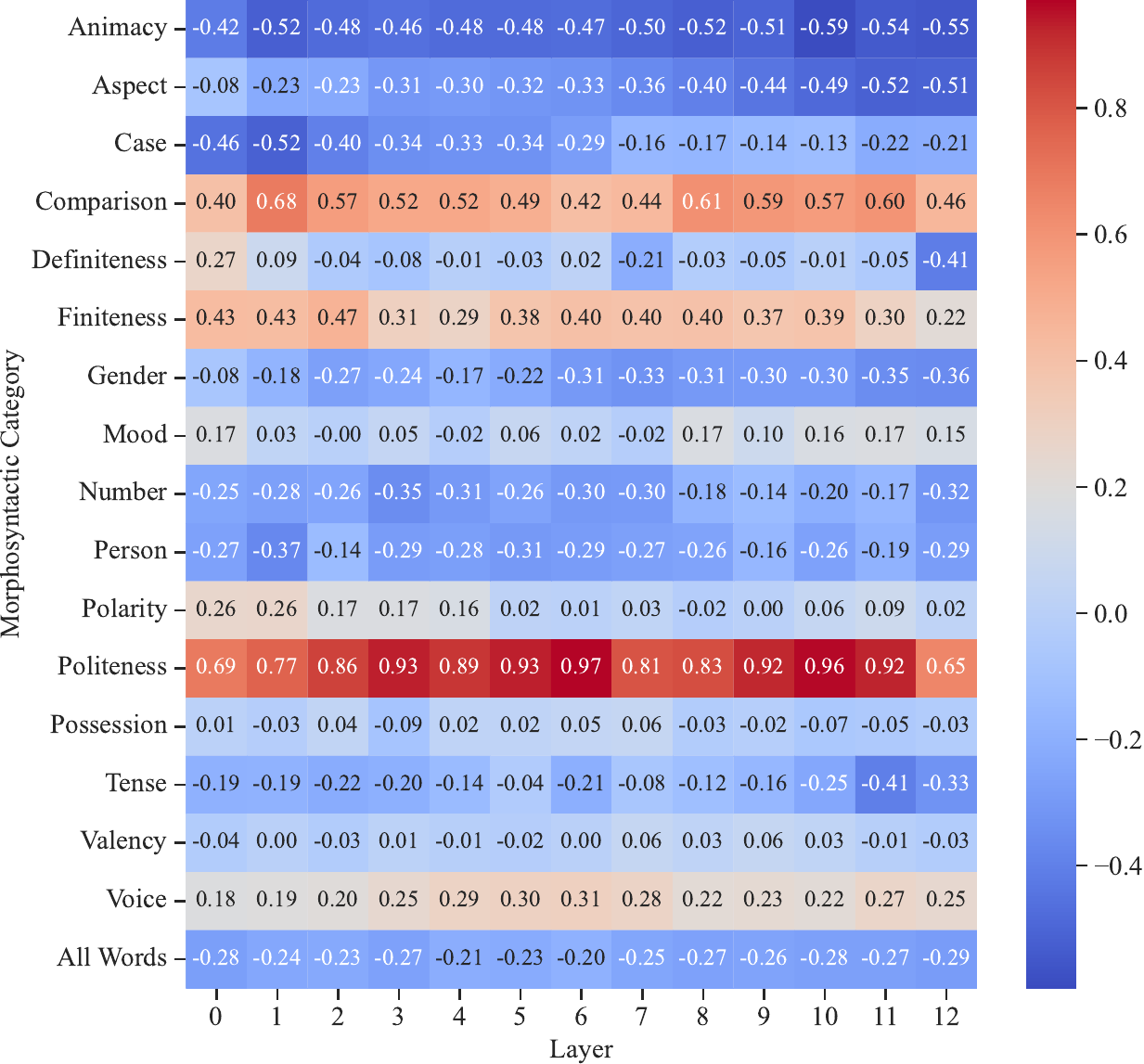}       
\caption{Pearson correlation results between the average $\mathrm{sig}(\ell)$ for all languages and their type-token ratio (TTR) for each morphosyntactic category among all layers.}
\label{fig:correlation_ttr_all_layers}
\end{figure}

We present the Pearson correlations between the average $\mathrm{sig}(\ell)$ for all languages and their number of unique characters for each morphosyntactic category among all layers in Figure~\ref{fig:correlation_uniq_chars_all_layers}. The results for type-token ratio (TTR) is presented in Figure~\ref{fig:correlation_ttr_all_layers}. Please note that in the figures, the last row is labeled as ``All Words'', representing the results obtained from using representations taken from the dataset with Part-of-Speech (PoS) attributes. Given that each lexical token in the dataset is associated with a PoS tag, this analysis encompasses the entire dataset, enabling us to observe and comprehend the overall trends captured in the representations.

\section{Additional Results for RQ2}
\label{app:rq2}
Adhering to the current standard practice of language sampling during pre-training of multilingual models, we also experimented a setting inspired by the approach described by \citet{Devlin_2019}. Following their approach, we applied a sampling technique to boost the representation of lower-resource languages. This involved sampling examples based on the probability $P(L) \propto |L|^{\alpha}$, where $P(L)$ represents the probability of selecting text from a given language during pre-training, and $|L|$ denotes the number of examples available in that language. For our study, we set the value of $\alpha$ to 0.3.

\section{Additional Results for RQ3}
\label{app:rq3}

We include in this section a phylogenetic tree analysis based on our approach, and a performance prediction experiment.

\subsection{Phylogenetic Tree}
\label{app:rq3-tree}

\begin{figure*}[ht]
   \centering
    \begin{subfigure}[b]{0.49\textwidth}
        \centering
         \includegraphics[width=\textwidth]{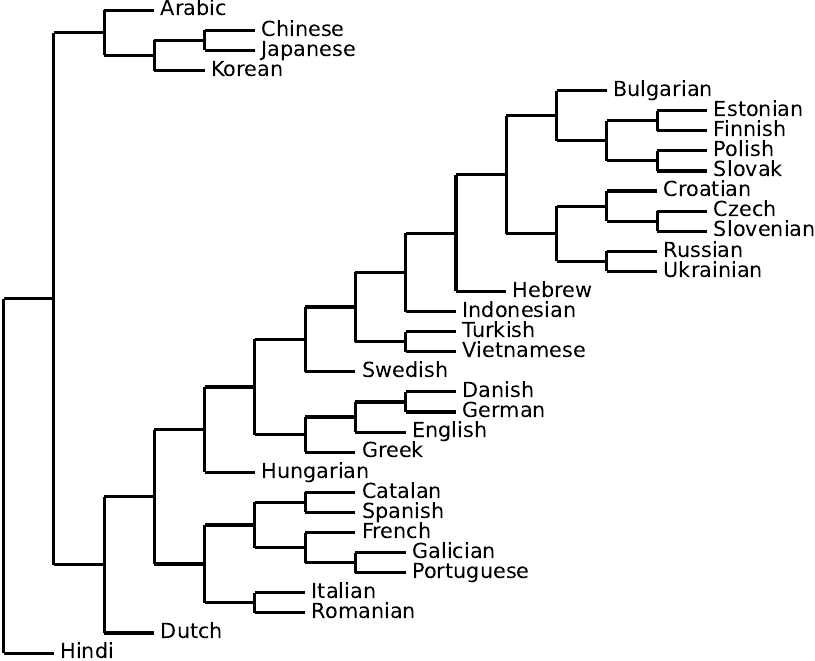}
         \caption{}
        \label{fig:tree-pos}
    \end{subfigure}
    \hfill
    \begin{subfigure}[b]{0.49\textwidth}
        \centering
         \includegraphics[width=\textwidth]{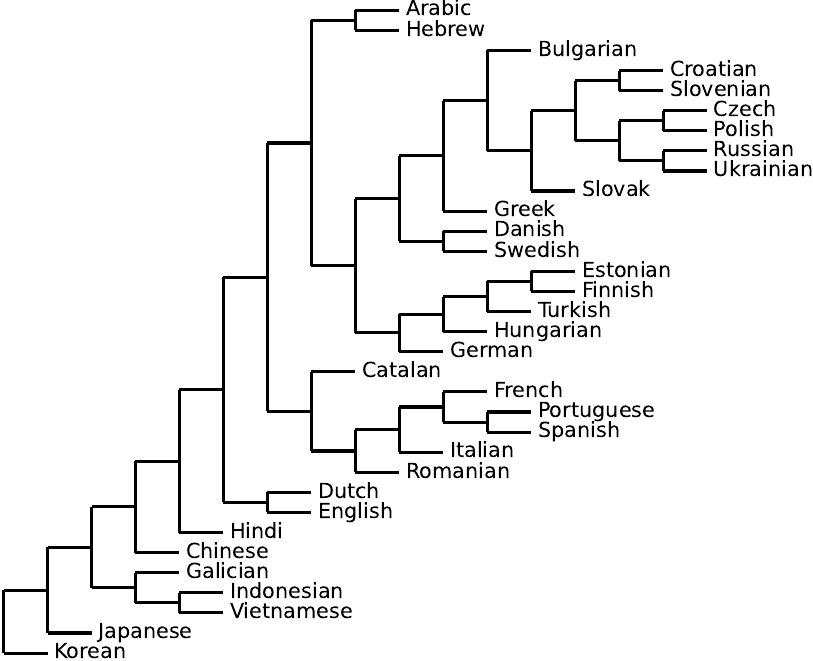}
         \caption{}
         \label{fig:tree-avg}
    \end{subfigure}
\caption{Phylogenetic trees of languages based on the distance between signatures vectors for all languages: (a) using overall representations; (b) averaged over morphosyntactic categories.}
\label{fig:tree}
\end{figure*}

We show the tree generated from layer 6's matrix is provided in Figure~\ref{fig:tree-pos}.

There is ongoing discussion over the specifics of the linguistic evolutionary phylogenetic tree of languages, and a tree model has limitations because not all evolutionary connections are fully hierarchical, and it is difficult to account for horizontal transmissions \citep{singh-etal-2019-bert}. Despite this, we can still see that the constructed phylogenetic tree closely matches the language tree that linguists created to describe the links and development of human languages. We can see that generally, Germanic, Romance, and Slavic languages are clustered in different sub-trees. In particular, West Slavic languages, South Slavic languages, and East Slavic languages are generally clustered together before being combined into the common Slavic language family. Also, Eastern Romance language Romanian are merged together with Western Romance languages to form the Romance language family cluster. Similar to the findings of \citet{singh-etal-2019-bert}, we also observe that trees generated across different layers are generally similar. They may have different structures as the branching of the tree may differ, but languages within the same family or genus are also close in the tree.  

So far, we have constructed trees based on the full slice of the data using representations from the PoS attribute. We also tried to generate trees using all other morphosyntactic attributes. However, since for most morphosyntactic attributes, some languages are always missing, i.e., the attribute is not available in that language, we construct an ``average'' tree by using the average distance matrices from all morphosyntactic categories except for PoS. If a language is missing in a category, we assign a distance of 1 to all other languages. We provide this average tree in Figure~\ref{fig:tree-avg}. In this average tree, we observe a better fit, for example, Arabic and Hebrew are now in the same branch, and Chinese and Japanese are in their own branch as they belong to language families that are distinct from all other languages.

\subsection{Performance Prediction}
\label{app:rq3-xtreme}

We present correlations for all layers for mBERT, XLM, XLM-R, and MMTE in Table~\ref{tab:correlation-xtreme-mbert-full}, Table~\ref{tab:correlation-xtreme-xlm-full}, Table~\ref{tab:correlation-xtreme-xlmr-full}, and Table~\ref{tab:correlation-xtreme-mmte-full}.

\begin{table*}[]
\small
\begin{tabular}{lccccccccc}
\Xhline{1pt}
\multirow{2}{*}{Layer} & \multicolumn{2}{c}{Pair sentence} & \multicolumn{2}{c}{Structured prediction} & \multicolumn{3}{c}{Question answering} & \multicolumn{2}{c}{Sentence retrieval} \\
                       & XNLI           & PAWS-X           & POS                 & NER                 & XQuAD     & MLQA      & TyDiQA-GoldP   & BUCC             & Tatoeba             \\
\Xhline{1pt}
                     & Acc.           & Acc.             & F1                  & F1                  & F1 / EM   & F1 / EM   & F1 / EM        & F1               & Acc.                \\
\Xhline{1pt}
0 & 0.01 & 0.71 & 0.26 & 0.47 & 0.33 / 0.15 & 0.03 / 0.15 & 0.61 / 0.20 & 0.64 & -0.11 \\
1 & -0.03 & 0.72 & 0.23 & 0.43 & 0.25 / -0.00 & -0.18 / -0.11 & 0.20 / -0.28 & 0.70 & 0.05 \\
2 & 0.01 & 0.76 & 0.32 & 0.49 & 0.31 / 0.09 & -0.15 / -0.04 & 0.36 / -0.16 & 0.68 & 0.05 \\
3 & 0.18 & 0.71 & 0.33 & 0.49 & 0.41 / 0.13 & -0.01 / 0.04 & 0.16 / -0.26 & 0.66 & 0.13 \\
4 & 0.10 & 0.71 & 0.38 & 0.55 & 0.43 / 0.19 & -0.01 / 0.10 & 0.41 / -0.12 & 0.65 & 0.12 \\
5 & 0.16 & 0.69 & 0.33 & 0.51 & 0.49 / 0.25 & 0.06 / 0.15 & 0.45 / -0.03 & 0.59 & 0.11 \\
6 & 0.24 & 0.74 & 0.41 & 0.59 & 0.54 / 0.28 & 0.12 / 0.20 & 0.54 / 0.03 & 0.66 & 0.18 \\
7 & 0.16 & 0.70 & 0.29 & 0.41 & 0.43 / 0.21 & 0.02 / 0.11 & 0.35 / -0.07 & 0.69 & 0.10 \\
8 & 0.09 & 0.56 & 0.14 & 0.20 & 0.32 / 0.12 & -0.11 / -0.04 & 0.13 / -0.18 & 0.64 & 0.03 \\
9 & 0.06 & 0.54 & 0.11 & 0.16 & 0.30 / 0.11 & -0.16 / -0.07 & 0.12 / -0.16 & 0.72 & -0.01 \\
10 & 0.15 & 0.55 & 0.15 & 0.23 & 0.37 / 0.15 & -0.02 / 0.04 & 0.14 / -0.13 & 0.71 & 0.01 \\
11 & 0.18 & 0.58 & 0.24 & 0.32 & 0.42 / 0.19 & 0.02 / 0.09 & 0.17 / -0.13 & 0.67 & 0.03 \\
12 & 0.36 & 0.67 & 0.36 & 0.46 & 0.60 / 0.35 & 0.23 / 0.31 & 0.41 / 0.05 & 0.72 & 0.15 \\
\Xhline{1pt}
\end{tabular}
\caption{Pearson correlations between $\mathrm{sig}(\ell)$ and XTREME benchmark performances of mBERT on various tasks.}
\label{tab:correlation-xtreme-mbert-full}
\end{table*}

\begin{table*}[]
\small
\begin{tabular}{lccccccccc}
\Xhline{1pt}
\multirow{2}{*}{Layer} & \multicolumn{2}{c}{Pair sentence} & \multicolumn{2}{c}{Structured prediction} & \multicolumn{3}{c}{Question answering} & \multicolumn{2}{c}{Sentence retrieval} \\
                       & XNLI           & PAWS-X           & POS                 & NER                 & XQuAD     & MLQA      & TyDiQA-GoldP   & BUCC             & Tatoeba             \\
\Xhline{1pt}
                     & Acc.           & Acc.             & F1                  & F1                  & F1 / EM   & F1 / EM   & F1 / EM        & F1               & Acc.                \\
\Xhline{1pt}
0 & -0.01 & 0.68 & 0.28 & 0.53 & 0.61 / 0.38 & 0.24 / 0.29 & 0.66 / 0.67 & 0.90 & -0.02 \\
1 & -0.11 & 0.71 & 0.25 & 0.52 & 0.51 / 0.22 & 0.03 / 0.06 & 0.59 / 0.53 & 0.96 & 0.13 \\
2 &  -0.07 & 0.74 & 0.34 & 0.57 & 0.54 / 0.27 & 0.05 / 0.11 & 0.65 / 0.60 & 0.97 & 0.12 \\
3 & 0.09 & 0.70 & 0.37 & 0.55 & 0.65 / 0.36 & 0.23 / 0.22 & 0.42 / 0.36 & 0.98 & 0.21 \\
4 & 0.05 & 0.70 & 0.39 & 0.59 & 0.68 / 0.40 & 0.21 / 0.27 & 0.58 / 0.54 & 0.94 & 0.20 \\
5 & 0.09 & 0.67 & 0.34 & 0.54 & 0.72 / 0.46 & 0.27 / 0.31 & 0.50 / 0.48 & 0.93 & 0.18 \\
6 & 0.17 & 0.72 & 0.41 & 0.63 & 0.75 / 0.48 & 0.32 / 0.35 & 0.67 / 0.65 & 0.94 & 0.27 \\
7 & 0.08 & 0.68 & 0.29 & 0.46 & 0.69 / 0.45 & 0.26 / 0.28 & 0.53 / 0.51 & 0.95 & 0.19 \\
8 & -0.02 & 0.54 & 0.16 & 0.24 & 0.55 / 0.34 & 0.12 / 0.13 & 0.29 / 0.26 & 0.97 & 0.09 \\
9 & -0.04 & 0.51 & 0.12 & 0.20 & 0.53 / 0.32 & 0.09 / 0.11 & 0.25 / 0.23 & 0.92 & 0.06 \\
10 & 0.07 & 0.52 & 0.17 & 0.26 & 0.61 / 0.38 & 0.23 / 0.22 & 0.29 / 0.28 & 0.92 & 0.08 \\
11 & 0.10 & 0.56 & 0.25 & 0.34 & 0.65 / 0.40 & 0.25 / 0.26 & 0.33 / 0.31 & 0.97 & 0.11 \\
12 & 0.30 & 0.65 & 0.36 & 0.46 & 0.81 / 0.56 & 0.46 / 0.48 & 0.43 / 0.43 & 0.96 & 0.24 \\
\Xhline{1pt}
\end{tabular}
\caption{Pearson correlations between $\mathrm{sig}(\ell)$ and XTREME benchmark performances of XLM on various tasks.}
\label{tab:correlation-xtreme-xlm-full}
\end{table*}

\begin{table*}[]
\small
\begin{tabular}{lccccccccc}
\Xhline{1pt}
\multirow{2}{*}{Layer} & \multicolumn{2}{c}{Pair sentence} & \multicolumn{2}{c}{Structured prediction} & \multicolumn{3}{c}{Question answering} & \multicolumn{2}{c}{Sentence retrieval} \\
                       & XNLI           & PAWS-X           & POS                 & NER                 & XQuAD     & MLQA      & TyDiQA-GoldP   & BUCC             & Tatoeba             \\
\Xhline{1pt}
                     & Acc.           & Acc.             & F1                  & F1                  & F1 / EM   & F1 / EM   & F1 / EM        & F1               & Acc.                \\
\Xhline{1pt}
0 & 0.11 & 0.80 & 0.67 & 0.62 & 0.63 / 0.34 & 0.59 / 0.62 & 0.58 / 0.52 & 0.91 & 0.14 \\
1 & -0.03 & 0.80 & 0.61 & 0.57 & 0.51 / 0.14 & 0.39 / 0.40 & 0.61 / 0.47 & 0.84 & 0.21 \\
2 & 0.02 & 0.83 & 0.67 & 0.64 & 0.58 / 0.24 & 0.46 / 0.50 & 0.63 / 0.47 & 0.85 & 0.23 \\
3 & 0.13 & 0.78 & 0.66 & 0.62 & 0.55 / 0.19 & 0.43 / 0.45 & 0.37 / 0.16 & 0.85 & 0.19 \\
4 & 0.15 & 0.80 & 0.73 & 0.68 & 0.71 / 0.39 & 0.56 / 0.61 & 0.59 / 0.46 & 0.90 & 0.31 \\
5 & 0.17 & 0.78 & 0.69 & 0.63 & 0.70 / 0.40 & 0.61 / 0.63 & 0.51 / 0.43 & 0.93 & 0.28 \\
6 & 0.24 & 0.82 & 0.73 & 0.70 & 0.73 / 0.42 & 0.63 / 0.64 & 0.68 / 0.62 & 0.89 & 0.35 \\
7 & 0.15 & 0.78 & 0.62 & 0.53 & 0.57 / 0.26 & 0.52 / 0.55 & 0.53 / 0.46 & 0.87 & 0.25 \\
8 & 0.04 & 0.66 & 0.47 & 0.33 & 0.36 / 0.05 & 0.32 / 0.37 & 0.30 / 0.23 & 0.88 & 0.13 \\
9 & 0.03 & 0.63 & 0.44 & 0.30 & 0.34 / 0.03 & 0.29 / 0.35 & 0.26 / 0.19 & 0.85 & 0.10 \\
10 & 0.13 & 0.64 & 0.49 & 0.36 & 0.41 / 0.10 & 0.37 / 0.41 & 0.29 / 0.24 & 0.86 & 0.11 \\
11 & 0.19 & 0.67 & 0.56 & 0.43 & 0.52 / 0.20 & 0.45 / 0.49 & 0.36 / 0.34 &0.85 & 0.15 \\
12 & 0.36 & 0.75 & 0.66 & 0.55 & 0.73 / 0.45 & 0.64 / 0.68 & 0.46 / 0.46 & 0.83 & 0.28 \\
\Xhline{1pt}
\end{tabular}
\caption{Pearson correlations between $\mathrm{sig}(\ell)$ and XTREME benchmark performances of XLM-R on various tasks.}
\label{tab:correlation-xtreme-xlmr-full}
\end{table*}

\begin{table*}[]
\small
\begin{tabular}{lccccccccc}
\Xhline{1pt}
\multirow{2}{*}{Layer} & \multicolumn{2}{c}{Pair sentence} & \multicolumn{2}{c}{Structured prediction} & \multicolumn{3}{c}{Question answering} & \multicolumn{2}{c}{Sentence retrieval} \\
                       & XNLI           & PAWS-X           & POS                 & NER                 & XQuAD     & MLQA      & TyDiQA-GoldP   & BUCC             & Tatoeba             \\
\Xhline{1pt}
                     & Acc.           & Acc.             & F1                  & F1                  & F1 / EM   & F1 / EM   & F1 / EM        & F1               & Acc.                \\
\Xhline{1pt}
0 & 0.02 & 0.73 & 0.33 & 0.51 & 0.51 / 0.53 & 0.06 / - & 0.92 / 0.35 & 0.59 & -\\
1 & -0.10 & 0.75 & 0.26 & 0.46 & 0.43 / 0.37 & -0.15 / - & 0.82 / 0.00 & 0.61 & -\\
2 & -0.06 & 0.78 & 0.38 & 0.53 & 0.48 / 0.44 & -0.14 / - & 0.93 / -0.04 & 0.58 & -\\
3 & 0.02 & 0.73 & 0.36 & 0.51 & 0.54 / 0.42 & 0.07 / - & 0.79 / -0.32 & 0.55 & -\\
4 & 0.01 & 0.74 & 0.44 & 0.58 & 0.59 / 0.52 & 0.00 / - & 0.91 / 0.06 & 0.58 & -\\
5 & 0.08 & 0.71 & 0.39 & 0.54 & 0.64 / 0.57 & 0.07 / - & 0.85 / 0.22 & 0.53 & -\\
6 & 0.15 & 0.76 & 0.44 & 0.61 & 0.69 / 0.59 & 0.13 / - & 0.91 / 0.31 & 0.58 & -\\
7 & 0.07 & 0.73 & 0.35 & 0.43 & 0.59 / 0.57 & 0.09 / - & 0.78 / 0.26 & 0.61 & -\\
8 & -0.02 & 0.59 & 0.21 & 0.23 & 0.46 / 0.44 & 0.00 / - & 0.59 / 0.14 & 0.55 & -\\
9 & -0.04 & 0.56 & 0.19 & 0.21 & 0.44 / 0.44 & -0.04 / - & 0.54 / 0.16 & 0.66 & -\\
10 & 0.05 & 0.58 & 0.20 & 0.27 & 0.51 / 0.46 & 0.10 / - & 0.55 / 0.20 & 0.64 & -\\
11 & 0.08 & 0.61 & 0.31 & 0.34 & 0.55 / 0.49 & 0.11 / - & 0.54 / 0.28 & 0.57 & -\\
12 & 0.21 & 0.69 & 0.40 & 0.46 & 0.72 / 0.61 & 0.28 / - & 0.66 / 0.45 & 0.63 & -\\
\Xhline{1pt}
\end{tabular}
\caption{Pearson correlations between $\mathrm{sig}(\ell)$ and XTREME benchmark performances of MMTE on various tasks.}
\label{tab:correlation-xtreme-mmte-full}
\end{table*}

\end{document}